\documentclass{article}

\PassOptionsToPackage{numbers, compress}{natbib}

\usepackage[dblblindworkshop,final]{neurips_2026}
\workshoptitle{World Models for High-Stakes Health: Reliable Clinical Trial
Simulation and Intervention-Aware Reasoning}

\usepackage[utf8]{inputenc}
\usepackage[T1]{fontenc}
\usepackage{amsmath,amssymb}
\usepackage{graphicx}
\graphicspath{{figures/}}
\usepackage{booktabs}
\usepackage{array}
\usepackage{longtable}
\usepackage{etoolbox}
\makeatletter
\patchcmd{\LT@makecaption}{\reset@font}{\reset@font\normalsize}{}{%
  \PackageError{wmhs}{Could not set longtable caption font}%
    {Check the longtable package definition.}}
\makeatother
\usepackage{float}
\usepackage[table]{xcolor}
\usepackage{tikz}
\usetikzlibrary{arrows.meta,positioning,fit,backgrounds,calc}
\usepackage{microtype}
\usepackage{url}
\usepackage[hidelinks]{hyperref}   %

\definecolor{chemo}{HTML}{B4462F}
\definecolor{radio}{HTML}{2E6E7E}
\definecolor{muted}{HTML}{6B7785}
\definecolor{soft}{HTML}{EEF1F2}
\newcommand{\best}[1]{\textbf{#1}}
\newcommand{\sd}[1]{\,\textcolor{muted}{$\scriptstyle\pm\mathrm{#1}$}}
\newcommand{\mk}[1]{\textcolor{#1}{\rule[0.02em]{0.42em}{0.42em}}\,}
\newcommand{\ourm}{\mk{radio}}
\newcommand{\theirm}{\mk{chemo}}
\newcommand{\nullm}{\mk{muted!55}}
\newcommand{\legend}{\\[2pt]{\scriptsize\color{muted}%
\ourm MGN \quad \theirm published model \quad \nullm plan-independent control}}

\newsavebox{\wmhsseedbox}
\newcommand{\wmhsseedtable}[1]{%
  \sbox{\wmhsseedbox}{\input{#1}}%
  \ifdim\wd\wmhsseedbox>\textwidth
    \resizebox{\textwidth}{!}{\usebox{\wmhsseedbox}}%
  \else\usebox{\wmhsseedbox}\fi}

\title{Beyond Low Error:\\
  Lessons from Tumor Growth Simulation}

\author{%
  Michał Zmysłowski \\
  Auton Lab, The Robotics Institute \\
  Carnegie Mellon University \\
  Pittsburgh, PA, USA \\
  \texttt{mzmyslow@andrew.cmu.edu} \\
  \And
  Abby Turner \\
  Auton Lab, The Robotics Institute \\
  Carnegie Mellon University \\
  Pittsburgh, PA, USA \\
  \texttt{abigailt@andrew.cmu.edu} \\
  \And
  Artur Dubrawski \\
  Auton Lab, The Robotics Institute \\
  Carnegie Mellon University \\
  Pittsburgh, PA, USA \\
  \texttt{awd@cs.cmu.edu} \\
}

\begin{document}

\maketitle

\begin{abstract}
\looseness=-1
A growing body of work uses a pharmacokinetic--pharmacodynamic tumor-growth simulator to train and evaluate models that predict tumor volume under different treatment plans.
In this work, we consider whether low error when predicting simulated tumors provides evidence that a model has learned how tumor volume changes with treatment.
We evaluate
thirteen published models, two controls that ignore the proposed plan, and Masked Growth Network
(MGN), a model matched to known properties of the simulated dynamics.
We compare evaluation with widely used sliding plans, which contain one treatment dose on one forecast day, with random plans that vary treatments independently across all forecast days.
We show that across data generation settings, proposed sliding plans account for only a limited share of variation in tumor outcomes. In the setting of our main results, proposed plans account for only \(5.5\%\) of outcome variation under sliding plans, compared with \(21.5\%\) under random plans.
With sliding treatment plans, persistence, which repeats the last observed tumor volume,
outperforms ten of thirteen published models. %
We also assess performance separately within deciles defined by tumor volume immediately before the proposed plan begins.
With random treatment plans, the largest-tumor decile's mean share of squared prediction
error ranges from $84.4\,\%$ to $94.4\,\%$ across MGN and the thirteen published models.
In each of the five smallest-tumor deciles, twelve of thirteen published models have higher mean contrast
error than a reference
returning the same tumor volume for every plan. In contrast, the model inspired by simulator dynamics improves on this reference across all deciles.
Its mean local dose responses increase with tumor size, although it overestimates the radiotherapy response.
Together, these results show that low prediction error alone does not establish that a model has learned how treatment affects tumor volume. Simulator-based evaluations should be interpreted in light of the treatment assignments observed during training, the plans proposed at evaluation, and the errors emphasized by the metric.
\end{abstract}

\section{Introduction}
\label{sec:intro}
Clinical decision support and trial emulation both require accurate predictions of how patient condition evolves under proposed treatment plans
\citep{schulam2017, hernan2016}.
While randomized trials can provide direct evidence about effects of possible treatment plans, they are not always ethical or feasible
\citep{hernan2016}. Thus a long line of work instead learns models to predict treatment outcomes from observational records
\citep{robins2000, lim2018, bica2020, li2021, melnychuk2022}.
Evaluating these models poses a fundamental challenge:
each patient record holds the outcome of the treatment plan actually administered, but not the outcomes that would have occurred under alternative plans \citep{holland1986}.
As such, evaluation with real-world datasets such as MIMIC \citep{johnson2016} can provide only limited evidence about model accuracy across treatment plans.

A common workaround is to train and evaluate models on simulated data. This enables us to sample realizations of patient outcomes under different proposed treatment plans. \citet{lim2018} used the pharmacokinetic--pharmacodynamic tumor-growth simulator of \citet{geng2017} to generate tumor volumes under alternative sequences of chemotherapy and radiotherapy. Subsequent studies have evaluated models on data generated by this simulator, using different evaluation settings \citep{bica2020, li2021, melnychuk2022, huang2024, kacprzyk2024,
xiong2024}.

In this work, we examine whether low prediction error on data from this simulator provides sufficient evidence that a model has learned how proposed treatment plans impact tumor volume.
We highlight two ways aggregate prediction error can conceal failures to predict treatment effects. First, different proposed plans may produce only small differences in resultant tumor volumes, allowing a model that ignores the plan to remain competitive (\S\ref{sec:sliding-vs-random}).
Second, cases in which tumor volume is largest at the start of a proposed plan may dominate the aggregate error. A model can thus achieve low error even when failing to predict how treatment will impact the volume of smaller tumors (\S\ref{sec:big-tumor}).

These failure modes call for close examination both of how informative proposed evaluation plans are and how model predictions respond to treatment. We quantify how much simulated outcomes vary across plans for the same history and supplement aggregate error with diagnostics that compare how model predictions and simulated outcomes respond to changes in treatment, reporting these comparisons across tumor sizes. Applied to thirteen published models, these analyses reveal discrepancies that aggregate rankings conceal, particularly for smaller tumors.
To show that aggregate accuracy and agreement under these finer-grained evaluations can be achieved together, we develop Masked Growth Network (MGN). Its architecture reflects two features of the underlying dynamics: tumor volume is updated multiplicatively, and treatments can affect only subsequent predictions. MGN achieves the lowest aggregate error in our main experiments and performs well on the finer-grained evaluations across tumor sizes. It therefore demonstrates that both forms of performance are attainable and illustrates how known properties of the data-generating process can inform design of an effective model.

\section{The simulator and evaluation settings}
\label{sec:benchmark}
\label{sec:setup}

We first describe the tumor-growth simulator, prediction task and evaluation settings.
The data consists of disjoint training, validation, and test cohorts of simulated patients with evolving tumors. For each patient, the simulator samples an initial tumor volume along with growth and treatment-response parameters. In conjunction with a time-series of treatment decisions, these determine the tumor's daily volume via a discrete-time growth equation \citep{geng2017, lim2018}.
We denote the tumor volume of patient $i$ on day $t$ by $V^i_t \in \mathbb{R}$.
Treatment decisions $a^i_t \in \{0,1\}^2$ indicate whether patient $i$ was given chemotherapy and radiotherapy on day $t$.

The prediction task requires a model to forecast how a patient's tumor volume will evolve under a proposed treatment plan, given the history of both volumes and treatment decisions. For patient $i$, the observed history ending on day $t$ forms a history \emph{window}, indexed by $(i,t)$. The history supplied to each model is specified in App.~\ref{app:models}. A model receives its history inputs along with a proposed treatment plan $a$ for the next $\tau=5$ days and predicts tumor volume on each of those days. For forecast day $k\in \{1,\ldots,\tau\}$, we write $V^i_{t+k}[a]$ for the simulator's volume under plan $a$ and
$\hat V^i_{t+k}[a]$ for the model's prediction.

In both training and test data, treatment decisions $a_t^i$ are
sampled with probabilities that depend on recent tumor history. %
During training, the model is passed history windows $(i,t)$ paired with the future treatments
observed in the simulated history. During evaluation, the model is passed each test window $(i,t)$ alongside a set of proposed future plans $\mathcal{A}^i_t$. Predictions under each plan are then compared with the corresponding simulated outcomes, generated from the same history and patient parameters.
We call each combination $(i,t,a)$ of history window $(i,t)$ and possible plan $a\in \mathcal{A}^i_t$ a \emph{window-plan pair}.

To measure performance at forecast day $k$, each predicted volume $\hat V^i_{t+k}[a]$ is compared with the corresponding simulated volume $V^i_{t+k}[a]$ and the resulting errors are aggregated across all window-plan pairs in the test data. %
We report
normalized root mean squared
error (nRMSE)
\citep{lim2018, bica2020, melnychuk2022}, which expresses RMSE as a percentage
of $v_{\max} \approx 1150\,\mathrm{cm^3}$,
the tumor volume at which the
simulator declares the patient dead. We also report mean absolute error (MAE), which
gives less weight to
large errors. The exact definitions
can be found in App.~\ref{app:metrics}.
Both scores are computed separately for each forecast day. We report them at the
fifth and final forecast day throughout.

Because nRMSE and MAE aggregate errors across plans, understanding the set of proposed treatment plans $\mathcal{A}^i_t$ is important for interpreting results. For example, if the proposed plans have little impact on tumor volume, models can score well without knowing anything about how treatment affects the volume. We next examine how treatments are assigned during training and which plans are proposed at evaluation.

\subsection{How is treatment assigned in training?}
\label{sec:eval-setup}

In our experiments, chemotherapy and radiotherapy are sampled independently given the
patient's history, with the same probability for both treatments. This probability depends
on the average tumor diameter over the preceding 15 days, with larger tumors making
treatment more likely.
A parameter $\gamma$ controls the strength of this dependence, yielding a family of treatment assignment strategies. For example, when $\gamma=0$ each treatment is assigned with probability $0.5$ regardless of tumor history. As $\gamma$ increases, probabilities become more sensitive to tumor size, making treatment increasingly predictable from tumor history.

This matters because $\gamma$ controls which combinations of tumor history and treatment are common in the training data. When treatment depends strongly on history, some combinations occur very rarely, leaving models with little data from which to learn how alternative treatments would impact tumor volume. Such history-dependent assignment is common in observational clinical data, where treatment decisions are closely informed by a patient's evolving condition.
We use $\gamma = 10$ in the main experiments,
the strongest dependence of treatment assignment on tumor history evaluated by
\citet{lim2018}, \citet{li2021} and \citet{huang2024}. %
We also report $\gamma = 4$ in App.~\ref{app:rosters}. This lies within the ranges
evaluated in the tumor experiments of \citet{melnychuk2022}
and the multi-step experiments of \citet{bica2020}.

\subsection{Which treatment plans are proposed at evaluation?}

At evaluation, the simulator
plays out each plan $a\in \mathcal{A}^i_t$ from the same history, patient parameters and future noise draw, so
the outcomes differ only by plan.
Two different types of $\mathcal{A}^i_t$ are considered, each
with $2\tau = 10$ plans to test per window. A \emph{sliding treatment plan} gives one
chemotherapy or radiotherapy dose on one of five forecast days, with no further treatment on any other days. %
A \emph{random treatment plan} assigns radiotherapy and chemotherapy using an independent
fair coin flip for each treatment on each day, leading to plans where one or both
treatments may occur on any given day.

While treatment assignments in training affect how readily models can learn treatment effects, evaluation plans $\mathcal{A}^i_t$ shape how informative low error is about that learning. Evaluating a model across a broad range of plans provides a more comprehensive assessment of what it has learned.

\section{The models}
\label{sec:model}
\label{sec:architecture}

We evaluate thirteen published models previously evaluated on data from this simulator,
either in their original papers or in later comparisons.
We also evaluate two controls that ignore the proposed treatment plan, and our own model Masked Growth Network (MGN).

\paragraph{Published models.} The published modeling strategies span a variety of approaches. We score thirteen models: MSM
\citep{robins2000}, RMSN \citep{lim2018}, G-Net \citep{li2021}, CRN \citep{bica2020}, CT and
EDCT \citep{melnychuk2022}, CT-ERM, CRN-ERM, OLS1 and OLS2 \citep{huang2024}, TARNet and CFR
\citep{shalit2017}, and INSITE \citep{kacprzyk2024}.
We use the tumor simulator and treatment-plan constructions
released with Causal Transformer \citep{melnychuk2022}. We compare models on the same
simulated cohorts and evaluation plans using the same scoring procedure (\S\ref{sec:setup}).
App.~\ref{app:models} describes each model and its history inputs, and summarizes exclusions.
App.~\ref{app:training} specifies the training and optimization settings for all models.
App.~\ref{app:prior-settings} compares the evaluation settings reported in prior studies.

\paragraph{Plan-independent controls.} The two
controls test what error is obtainable by methods that do not depend on the treatment plan $a\in \mathcal{A}^i_t$ when forecasting. Since they do not take the proposed plan as input,
predicted volumes for a given window
are the same under every plan. The first control is a persistence model, which repeats the last observed volume $V^i_t$ for
the whole horizon. The second control is a constant model, predicting the mean volume from the training set for every window and forecast day.

\paragraph{Masked Growth Network.} %
MGN is designed around two properties of the simulated tumor dynamics.
First, it predicts tumor volume as a multiplicative change
from the last observed volume, reflecting the simulator's volume update.
Second, its masked head enforces temporal
order: the forecast for day $t+k$ is a function only of the first $k$ treatment decisions in the proposed plan.
App.~\ref{app:mgn} gives more details about the architecture.
App.~\ref{app:rosters} compares variants with and
without the mask and with different output compositions.

\section{Results}
\label{sec:results}

The following table reports results at the five-day horizon for $\gamma=10$, where treatment assignment depends most strongly on tumor history, under sliding and random proposed plans.

\begin{table}[H]\centering\small
\setlength{\tabcolsep}{3.5pt}
\caption{Prediction errors at $\gamma = 10$, $\tau = 5$. Lower is better and \textbf{bold} is the best in each column.
Entries are the mean $\pm$ sample standard deviation
over ten seeds; each seed fixes both the initialization and an independently simulated cohort.
Citations are in App.~\ref{app:models}; the five MGN ablations and the $\gamma = 4$ comparison
are in App.~\ref{app:rosters}.}
\label{tab:wmhs-g10}
\begin{tabular}{l rr @{\qquad} l rr}
\toprule
\multicolumn{3}{c}{\textbf{sliding treatment plans}} & \multicolumn{3}{c}{\textbf{random treatment plans}} \\
\cmidrule(lr){1-3}\cmidrule(lr){4-6}
model & nRMSE \% & MAE & model & nRMSE \% & MAE \\
\midrule
\ourm MGN & \best{3.188}\sd{0.81} & \best{10.250}\sd{4.4}  & \ourm MGN & \best{2.507}\sd{0.55} & \best{8.482}\sd{3}  \\
\theirm TARNet & 3.497\sd{0.85} & 12.059\sd{4.2}  & \theirm TARNet & 2.788\sd{0.65} & 11.490\sd{3.3}  \\
\theirm CFR & 3.853\sd{1.1} & 14.286\sd{5.5}  & \theirm CFR & 2.934\sd{0.82} & 11.802\sd{3.3}  \\
\theirm INSITE & 5.004\sd{1.2} & 16.964\sd{5.1}  & \theirm CRN-ERM & 3.072\sd{1.00} & 11.909\sd{7.3}  \\
\nullm persistence & 5.128\sd{1.5} & 17.295\sd{7.3}  & \theirm INSITE & 3.077\sd{0.68} & 15.236\sd{3.2}  \\
\theirm CT-ERM & 6.112\sd{1.8} & 30.051\sd{5.7}  & \theirm CT-ERM & 3.153\sd{0.64} & 14.442\sd{2.2}  \\
\theirm MSM & 6.736\sd{1.9} & 25.497\sd{11}  & \theirm CT & 3.268\sd{1.0} & 15.974\sd{12}  \\
\theirm CT & 7.146\sd{2.5} & 33.298\sd{21}  & \theirm G-Net & 3.296\sd{0.69} & 12.284\sd{3.2}  \\
\theirm G-Net & 7.518\sd{2.6} & 29.008\sd{12}  & \theirm OLS1 & 3.564\sd{0.81} & 15.398\sd{3.4}  \\
\theirm CRN-ERM & 7.984\sd{2.6} & 31.800\sd{19}  & \theirm OLS2 & 3.673\sd{0.85} & 17.716\sd{3.7}  \\
\theirm OLS2 & 8.060\sd{2.1} & 35.121\sd{12}  & \theirm CRN & 3.823\sd{0.98} & 13.583\sd{4.3}  \\
\theirm CRN & 10.599\sd{3.0} & 44.645\sd{19}  & \theirm EDCT & 3.926\sd{0.88} & 13.515\sd{4.2}  \\
\theirm EDCT & 10.714\sd{2.7} & 42.419\sd{15}  & \nullm train mean & 4.640\sd{1.1} & 21.972\sd{4.4}  \\
\theirm OLS1 & 10.761\sd{2.8} & 49.012\sd{17}  & \nullm persistence & 12.713\sd{3.5} & 55.454\sd{22}  \\
\nullm train mean & 12.539\sd{3.3} & 58.851\sd{20}  & \theirm MSM & 14.493\sd{4.0} & 63.741\sd{25}  \\
\theirm RMSN & 20.105\sd{9.9} & 87.664\sd{60}  & \theirm RMSN & 25.420\sd{13} & 102.924\sd{76}  \\
\bottomrule
\end{tabular}
\legend
\end{table}
Under both plan constructions, MGN has the lowest nRMSE and MAE.
The appendix ablations identify multiplicative output as a key driver of this performance.
Replacing it with either of the two additive variants tested there increases both errors
under both plan constructions, with or without temporal masking (App.~\ref{app:rosters}).

Under sliding treatment plans,
which each contain only one treatment across forecast days,
persistence ranks fifth of sixteen by mean nRMSE,
ahead of ten of the thirteen published models. With random treatment plans, which may contain multiple treatments across days,
persistence
is outperformed by all methods except for MSM and RMSN. Its ranking
therefore changes sharply between plan constructions.
We next examine a feature of proposed plans $\mathcal{A}^i_t$ that may help explain this discrepancy.

\subsection{How much does tumor volume vary across plans?}
\label{sec:sliding-vs-random}

The persistence model falls sharply in ranking when moving from sliding to random evaluations, suggesting that the proposed plans are more important for forecasting in the latter setting.
To quantify this, we consider the \emph{plan-controlled variance share} defined by
\[
  s \;=\; \frac{\sum_{i,t} \sum_{a \,\in\, \mathcal{A}^i_t}
                \big( V^i_{t+\tau}[a] - \bar V^i_t \big)^2}
               {\sum_{i,t} \sum_{a \,\in\, \mathcal{A}^i_t}
                \big( V^i_{t+\tau}[a] - \bar V \big)^2},
\]
where $\bar V^i_t$ is the mean final-step tumor volume across the plans
proposed for window $(i,t)$, and $\bar V$ is the mean across all window-plan pairs.
The numerator measures how much outcomes vary across proposed plans within
each window, holding the patient, history and future noise draw fixed.
The denominator measures how much outcomes vary across all windows and
plans. Thus, $s$ is the fraction of total outcome variation that comes
from changing the proposed plan within a window.

\begin{figure}[H]
  \centering
  \includegraphics[width=0.65\textwidth]{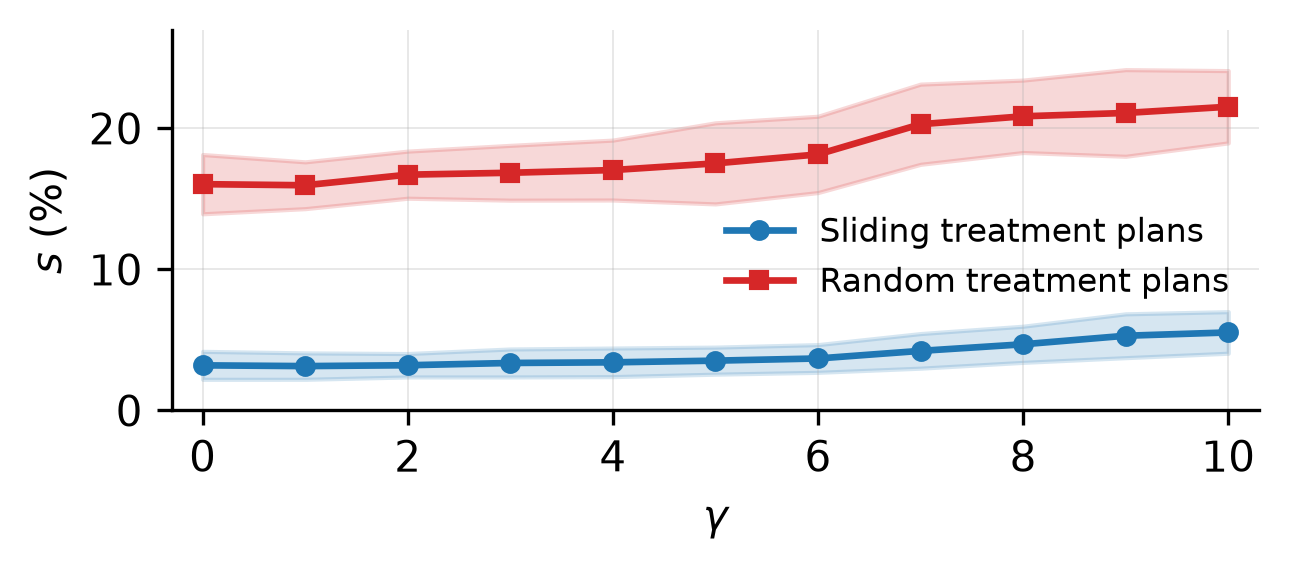}
  \caption{The plan-controlled variance share $s$ at the five-day horizon.
    We compute the share separately for ten seeds and then report its mean; shading shows one
    sample standard deviation on either side. Each seed contains approximately $50{,}000$
    windows with finite fifth-day outcomes under all ten plans.
    This share is larger with random treatment plans than with sliding
    treatment plans throughout the evaluated range of treatment-assignment
    strength $\gamma$.}
  \label{fig:plan-variance}
\end{figure}

When training treatment depends strongly on tumor history ($\gamma = 10$),
the plan-controlled variance share $s$ is $5.5\,\%$ with sliding treatment
plans, which contain just one dose each. With random treatment plans,
which vary each treatment independently at every forecast step,
the mean share is $21.5\,\%$. Random treatment plans therefore control a larger
fraction of total outcome variation than sliding treatment plans.
Figure~\ref{fig:plan-variance} shows that this ordering holds throughout
the evaluated $\gamma$ sweep.
The small plan-controlled variance share under sliding treatment plans
is consistent with persistence remaining competitive without using
the proposed plan (Table~\ref{tab:wmhs-g10}).
We therefore use random treatment plans for our investigation in \S\ref{sec:do-models-learn}.

\section{Have models learned how treatment changes tumor volume?}
\label{sec:do-models-learn}
For the remainder of our analysis, we focus on results assuming random treatment plans. %
Compared with sliding plans, random plans control a larger fraction of total outcome variation, and persistence is no longer competitive (\S\ref{sec:sliding-vs-random}).
Despite these advantages, in this section we show that low aggregate error on random plans can still conceal failures to capture how treatment changes tumor volume. %

First we show that the aggregate prediction error used
is primarily driven by tumors that are largest when the proposed plan begins (\S\ref{sec:big-tumor}). We therefore report two complementary diagnostics
separately across tumor size.
The first of these, which we call contrast error, assesses whether the models' predictions reflect how simulated outcomes differ between plans (\S\ref{sec:contrast-error}).
A second local-response diagnostic compares the
direction and magnitude of each model's local dose response with the
simulator's around a no-treatment plan (\S\ref{sec:jacobian}). %

\subsection{Where does the error concentrate?}
\label{sec:big-tumor}
In this section, we consider which tumor sizes dominate the squared error underlying nRMSE for MGN and the published models.
This matters because an aggregate score dominated by large tumors may conceal disagreement
between the treatment effects learned by the models and those generated by the simulator for
smaller tumors.

We group test windows into deciles by the last observed tumor volume before the proposed plan begins.
Across MGN and the thirteen published models, the largest tumor-volume decile contributes
$84.4\,\%$--$94.4\,\%$ of total squared error on average over ten seeds, despite containing
approximately $10\,\%$ of the window-plan pairs. App.~\ref{app:error-concentration}
reports each model's mean and 95\% confidence interval.
Therefore, we report the next two treatment-effect diagnostics separately for each tumor-volume decile.

\subsection{Do the models' predictions reflect how simulated outcomes differ between plans?}
\label{sec:contrast-error}

For each history window, the sum of squared prediction errors across its proposed treatment
plans decomposes exactly into what we call a level term and a contrast term (App.~\ref{app:contrast-split}).
The level term compares the
mean prediction across plans with the mean simulated volume. A model can predict
this mean correctly while its predictions fail to reflect how treatment plans change tumor volume.

We therefore focus on the contrast term, which compares centered predicted volumes with centered
simulated volumes. Each centered volume shows how far the volume under a particular plan lies
above or below the mean across plans. The comparison thus focuses on differences in tumor
volume between plans. Within each history window, these differences arise solely from changing
the treatment plan. This comparison therefore targets errors in the treatment effects reflected
in the model's predictions.

More precisely, for a set $B$ of window-plan pairs $(i,t,a)$ with $a \in \mathcal{A}^i_t$, we define
\emph{contrast error} by
\begin{equation}
\label{eq:contrast-error}
  E(B) \;=\; \sqrt{ \frac{1}{|B|} \sum_{(i,t,a) \,\in\, B}
      \Big( \big( \hat V^i_{t+\tau}[a] - \bar{\hat V}^i_t \big)
          - \big( V^i_{t+\tau}[a] - \bar V^i_t \big) \Big)^2 } ,
\end{equation}
where $\bar{\hat V}^i_t$ and $\bar V^i_t$ are the means of the predicted and simulated
volumes at horizon $\tau$, respectively, across the proposed plans in $\mathcal{A}^i_t$.
We report contrast error in $\mathrm{cm^3}$.

To interpret these scores, we compare them with the contrast error of predictors that ignore
the proposed treatment plan. Persistence and the training mean both give the same prediction
under every plan for a given history window. Their centered predictions are therefore zero.
For any such predictor, Eq.~\ref{eq:contrast-error} reduces to
\begin{equation}
\label{eq:contrast-null}
  E_0(B) \;=\; \sqrt{ \frac{1}{|B|} \sum_{(i,t,a) \in B}
      \big( V^i_{t+\tau}[a] - \bar V^i_t \big)^2 } .
\end{equation}
We call this shared value the \emph{plan-independent reference} $E_0$.

\begin{figure}[H]
  \centering
  \includegraphics[width=0.78\textwidth]{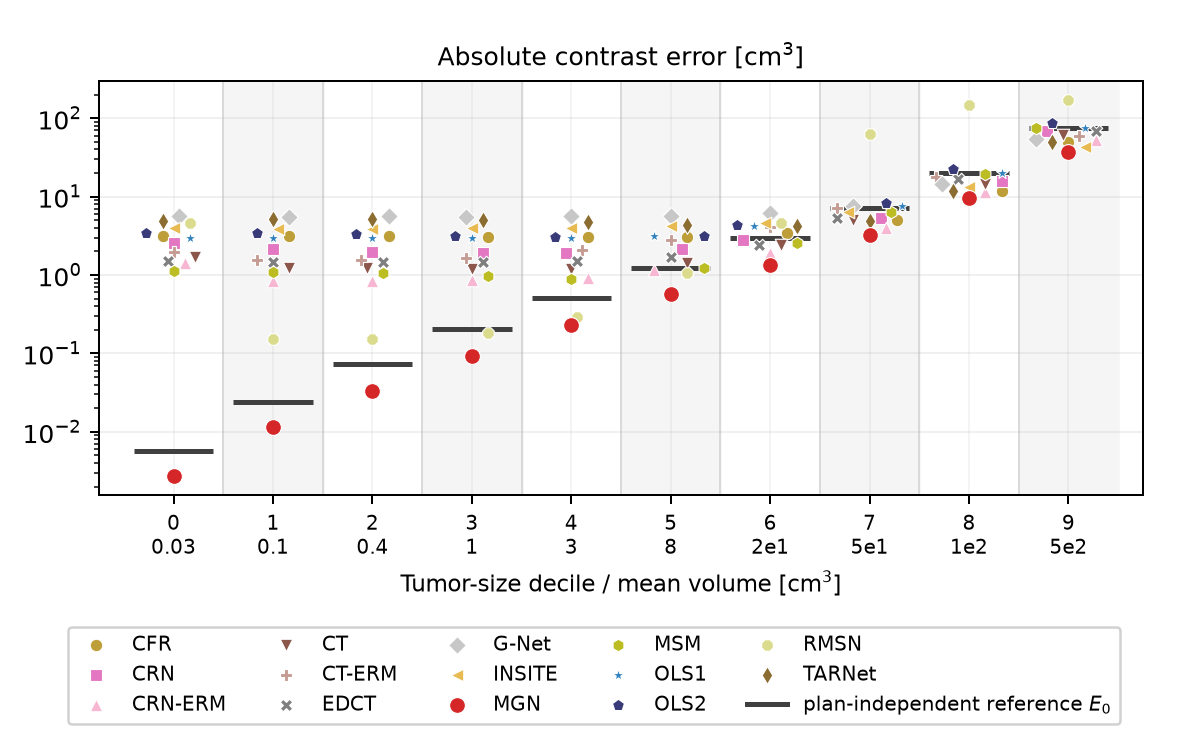}
  \caption{\textbf{Absolute contrast error.} Random treatment plans, $\gamma = 10$,
  $\tau = 5$. Points and reference segments show equally weighted means over ten base seeds
  of absolute contrast error in $\mathrm{cm^3}$ by tumor-size decile.
  Below each decile index, the second x-axis row gives the ten-seed mean last observed tumor volume in
  that decile, in $\mathrm{cm^3}$.
  Points are displaced horizontally within each decile to prevent overlap. The black segment
  marks the plan-independent reference $E_0$ (Eq.~\ref{eq:contrast-null}); points above it have
  greater contrast error than a plan-independent predictor. The vertical axis is logarithmic.
  App.~\ref{app:contrast-cells} adds the MGN ablations and reports the other evaluation settings.}
  \label{fig:contrast-by-size}
\end{figure}

Figure~\ref{fig:contrast-by-size} reports contrast error separately for each tumor-volume
decile, defined by the last observed tumor volume as in \S\ref{sec:big-tumor}.
For each decile, $B$ contains its window-plan pairs. The black segment marks the
plan-independent reference. Points and reference segments show means over ten base seeds.
Additional contrast-error results appear in App.~\ref{app:contrast-cells}.

Over ten seeds, twelve of thirteen published models have higher mean contrast error than the
plan-independent reference in each of the five smallest deciles. RMSN has lower means in deciles
3 and 4, but its paired 95\% intervals for the difference from the reference include zero:
the apparent improvement varies across seeds (App.~\ref{app:contrast-cells}).
MGN is below the reference in every decile for every seed; all ten paired intervals lie below zero.
Its performance across sizes is consistent with the multiplicative output of
\S\ref{sec:architecture}, which incorporates scaling by the last observed tumor
volume directly.

Contrast error alone does not show
whether the treatment effect learned by the model underestimates, exaggerates or reverses the
simulator's treatment effect. To distinguish these
behaviors locally, we next compare the direction and magnitude of each model's local dose response
with the simulator's around a no-treatment plan.

\subsection{Do the predictions reflect the simulator's local dose response?}
\label{sec:jacobian}
In this section, we examine the response to increasing each treatment dose away from zero. %
This offers a strategy to assess the direction and magnitude of the local dose response reflected in the model's predictions separately for chemotherapy and radiotherapy. %

\begin{figure}[H]
  \centering
  \includegraphics[width=0.78\textwidth]{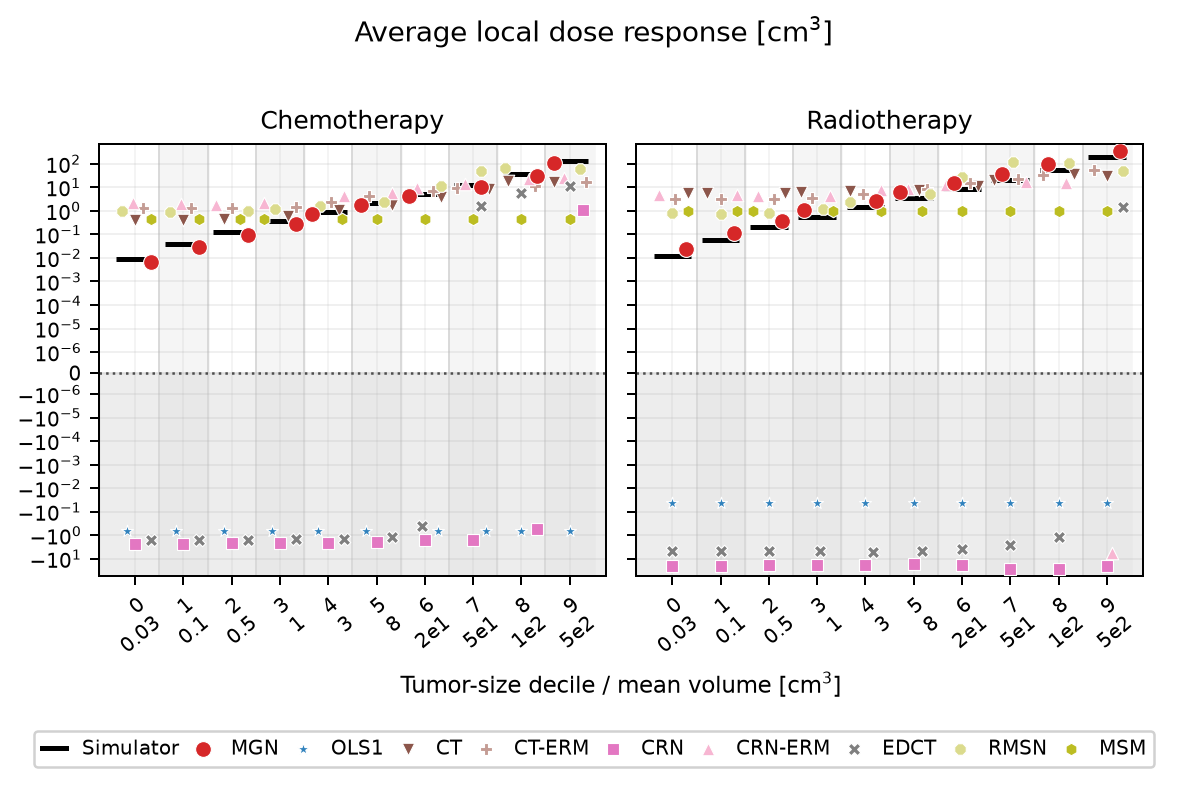}
  \caption{\textbf{Average local dose response.} Random treatment plans, $\gamma = 10$,
    $\tau = 5$. Each point shows the equally weighted mean over ten base seeds of the
    mean local dose response over windows within one tumor-size decile, in $\mathrm{cm^3}$.
    Points are displaced horizontally within
    each decile to prevent overlap. Below each decile index, the second x-axis row gives the mean
    last observed tumor volume in that decile, averaged over ten seeds, in $\mathrm{cm^3}$.
    The black segment marks the simulator's response averaged over the same ten seeds. Positive
    values indicate tumor reduction; the shaded negative region indicates tumor increase.
    The vertical axis uses a symmetric logarithmic scale.
    App.~\ref{app:jacobian-method} describes the
    computation, and App.~\ref{app:jacobian-cells} reports the other evaluation settings.}
  \label{fig:jacobian-grid}
\end{figure}

To do so, we relax the binary treatment indicators $a^i_t \in \{0,1\}^2$ to continuous values
$u^i_t \in [0,1]^2$, where 0 indicates no treatment and one indicates a full dose. Of note, although continuous treatment values are permitted by the underlying tumor-growth simulator, models we consider here train and evaluate with only simplified binary treatments in $\{0,1\}$. The derivatives therefore characterize the continuous response implied by each model, rather than one learned from intermediate-dose examples.
Writing $r$
for chemotherapy or radiotherapy, we summarize this local dose response as 
\[
  J^i_t(r) =
  -\frac{1}{\tau}\sum_{j=1}^{\tau}
  \left.
  \frac{\partial \hat V^i_{t+\tau}}
       {\partial u^i_{r,\,t+j}}
  \right|_{u=0}.
\]
For each window and treatment, $J^i_t(r)$ averages the negated
final-step derivatives over the five forecast days, in $\mathrm{cm^3}$.
We negate the derivatives so that positive values indicate tumor reduction and negative
values indicate tumor increase.
We then average these values over windows within each tumor-volume decile.
Comparison with the simulator's corresponding
average shows whether the predicted local tumor reduction is underestimated or exaggerated.
Figure~\ref{fig:jacobian-grid} plots the average local dose response by tumor-volume decile
against the simulator's own derivative, which is available in closed form. Both are averaged
over ten base seeds. Additional local dose-response results appear in App.~\ref{app:jacobian-cells}.
App.~\ref{app:jacobian-method} describes how the derivatives are computed.

The simulator's local dose response grows approximately in proportion to tumor volume. The
published models' ten-seed means show different patterns across sizes. Some approach the simulator for larger
tumors, but their response magnitudes on the smallest tumors remain orders of magnitude too
large. No displayed published model matches both of the simulator's local dose responses across
all ten deciles. MGN's mean responses retain the increase with tumor size. In the main setting,
its chemotherapy means are below the simulator's and its radiotherapy means are above it
in every decile (App.~\ref{app:jacobian-cells}).

This diagnostic is local because every derivative averaged in $J^i_t(r)$ is evaluated at no
treatment. It summarizes the response to small dose changes rather than the finite effect of
administering a full dose. %
\section{Discussion}
\label{sec:discussion}

Low aggregate prediction error is insufficient evidence that a model has learned how treatment
changes tumor volume. %
With sliding treatment plans, the proposed plans control
little of the outcome variation, and persistence remains competitive without using the plan.
Random treatment plans increase the plan-controlled
variance share and make persistence less competitive
(\S\ref{sec:sliding-vs-random}). When evaluating models on simulated data, we should therefore
consider how much outcome variation the proposed plans control when selecting evaluation plans and interpreting aggregate prediction error. For example, this can help identify evaluation setups in which models can score well by predicting the mean outcome across plans.

Changing the proposed plans addresses only part of the problem. Even under
random treatment plans, twelve of thirteen published models have higher mean contrast error than the
plan-independent reference in each of the five smallest tumor-volume deciles; RMSN's lower
means in deciles 3 and 4 have paired confidence intervals that include zero
(\S\ref{sec:contrast-error}). Aggregate squared error concentrates in the
largest tumors, obscuring these failures on smaller ones
(\S\ref{sec:big-tumor}). We therefore recommend reporting treatment-effect
diagnostics separately from aggregate outcome error over possible plans. When possible given the simulator and model, comparing derivatives can further help examine whether direction and magnitude of the local dose response reflected in the model matches the underlying simulation.
Reporting these diagnostics by tumor-volume
decile reveals whether agreement with the simulator extends beyond the largest tumors, which
dominate the aggregate score.

Our model MGN serves to demonstrate that by constructing architectures reflecting underlying properties of the dynamics, models may avoid performance pitfalls that are possible in spite of low aggregate error. %
Its masked head enforces temporal order, and
its multiplicative output incorporates the simulator's scaling of treatment
effects with tumor volume. With random treatment plans,
it improves on the plan-independent reference across all tumor-size deciles.
Its mean local responses increase with tumor size, but their magnitudes do not consistently
match the simulator: radiotherapy is overestimated across all deciles in the main setting.
The diagnostics therefore expose discrepancies even for a model with strong contrast performance.
Future experiments could %
explore alternative models of how treatment affects tumor
growth and test whether the treatment effects learned by MGN still agree
with the modified simulator's treatment effect.

\section{Responsible-use statement}
\label{sec:responsible-use}

\paragraph{Limitations.} Every number in this paper comes from one simulator with counterfactual outcomes that are exact by construction. The
proposed plans follow simplified rules for model evaluation rather than clinically motivated
treatment schedules. No model here,
ours included, is evaluated on real patients, and ours is built to the simulator's structure.

\paragraph{Uncertainty.}
Aggregate nRMSE and MAE results and the plan-variance sweep are means over ten seeds, accompanied
by sample standard deviations. Error concentration, contrast error and local dose response
are also evaluated over ten seeds. Contrast-error and local-response figures show ten-seed
means, and their compact appendix tables report means and sample standard deviations for
three decile ranges. The error-concentration table and paired contrast comparisons report
95\% Student-$t$ intervals for the across-seed mean under repeated cohort generation and fitting.
The seed is the replication unit throughout.

\paragraph{Clinical and societal impacts.} The risk is treating aggregate error as evidence that
a model has learned treatment effects. Models can rank well by aggregate error while their
contrast error on smaller tumors remains no better than the plan-independent reference. No model
evaluated here is ready to inform treatment.

\paragraph{Mitigations.} Report the plan-controlled variance share and contrast error by
tumor-volume decile. Alongside aggregate error, compare the local dose response reflected in
predictions with the simulator's. Treat performance on simulated data as a reason for further testing,
not clinical validation.

\bibliographystyle{plainnat}
\bibliography{references}

\clearpage
\appendix

\section{Prediction-error metrics}
\label{app:metrics}

Let $N = \sum_{i,t}|\mathcal{A}^i_t|$ be the total number of window-plan pairs. At forecast day
$k$, normalized root mean squared error is
\begin{equation}
  \operatorname{nRMSE}(k)
  = \frac{100}{v_{\max}}
    \sqrt{\frac{1}{N}\sum_{i,t}\sum_{a\in\mathcal{A}^i_t}
      \big(\hat V^i_{t+k}[a] - V^i_{t+k}[a]\big)^2}.
  \label{eq:nrmse}
\end{equation}
Mean absolute error is
\begin{equation}
  \operatorname{MAE}(k)
  = \frac{1}{N}\sum_{i,t}\sum_{a\in\mathcal{A}^i_t}
      \big|\hat V^i_{t+k}[a] - V^i_{t+k}[a]\big|.
  \label{eq:mae}
\end{equation}

\section{What the published models do}
\label{app:models}

The thirteen published models use different approaches to predicting
tumor volume under proposed treatment plans. Below, we describe how each
model works and how it handles history-dependent treatment assignment,
then explain why other models were not included. We fit these implementations on the
same simulated training cohort and score them on the same evaluation plans
(\S\ref{sec:setup}). App.~\ref{app:prior-settings} distinguishes these settings from
those reported in earlier studies.

\paragraph{Implementation sources.}
We use the implementations of MSM, RMSN, G-Net, CRN, CT and EDCT distributed with
Causal Transformer \citep{melnychuk2022}. CT-ERM and CRN-ERM use the corresponding
implementations with balancing disabled. INSITE uses its authors' released
implementation \citep{kacprzyk2024}. TARNet, CFR, OLS1, OLS2 and MGN use our
implementations.

\paragraph{Weighting and trajectory simulation.}
Marginal Structural Models (MSM) \citep{robins2000} estimate treatment
effects by fitting an outcome model with inverse-probability-of-treatment
weights to account for history-dependent assignment. The implementation
used here estimates treatment probabilities with logistic regressions and
predicts tumor volume with weighted linear regressions.
Recurrent Marginal Structural Networks (RMSN) \citep{lim2018} use the same
weighting principle, with recurrent networks estimating treatment
probabilities and an encoder--decoder network predicting outcomes under
the proposed plan.
G-Net \citep{li2021} uses recurrent networks within g-computation to model
the conditional evolution of outcomes and covariates. It simulates
trajectories under the proposed plan and averages their outcomes to
estimate the expected tumor volume.

\paragraph{Representation balancing and its ablations.}
These methods aim to reduce differences between treatment groups in the
learned representation while preserving information needed to predict
outcomes. The Counterfactual Recurrent Network (CRN) \citep{bica2020}
uses an LSTM encoder--decoder. The Causal Transformer (CT)
\citep{melnychuk2022} exchanges information between separate input
streams through cross-attention; in our evaluation, these streams
contain treatments and tumor volumes. Its encoder--decoder variant,
EDCT, combines the inputs within each time step and uses a decoder
that attends to the encoded history.

For CRN, CT and EDCT, we use the domain-confusion variants released by
\citet{melnychuk2022}. Their training alternates between fitting a
treatment classifier and learning representations that encourage its
predictions towards a uniform distribution, alongside outcome
prediction. CT-ERM and CRN-ERM disable this balancing objective and
retain the mean squared outcome-prediction loss, following the ablation
studied by \citet{huang2024}.

TARNet and CFR \citep{shalit2017} were developed for a static binary
treatment. Both use a shared representation network and separate
outcome heads for the treatment groups; CFR adds a penalty for
differences between their representation distributions. We implement
both for our prediction task using flattened histories, the forecast step
and the proposed treatments up to that step as inputs. Four outcome
heads correspond to neither treatment, chemotherapy, radiotherapy and
both, with the head selected by the treatment at that step. Our CFR
implementation penalizes differences between each group's mean
representation and the pooled mean; TARNet omits this penalty.

\paragraph{No correction.} OLS1 fits a pooled linear regression with
treatment as an input; OLS2 fits separate regressions for treatment
groups \citep{shalit2017,huang2024}. We adapt both to our prediction task by
fitting each forecast day separately using the observed history and
only that day's treatment from the proposed plan.

\paragraph{Equation discovery.} INSITE \citep{kacprzyk2024} uses sparse
regression to fit an ordinary differential equation for each treatment
combination. It first fits population equations, then adjusts their
coefficients to each patient's observed history. It predicts tumor volume
by integrating these equations under the proposed plan. We use the authors'
released implementation with its polynomial library of candidate functions.

\paragraph{History inputs.}
We retain the history inputs used by the released implementations. For MGN and our
implementations of TARNet, CFR, OLS1 and OLS2, we use a shared 16-day history window.
This covers the 15-day history used by the simulator to assign treatments. Shorter
histories are padded so that these models can be evaluated at the same forecast
origins as the released models. Table~\ref{tab:history-inputs} specifies the history
inputs for every model.

\begin{table}[H]
  \centering
  \small
  \caption{Observed history used by each implementation in our comparison.}
  \label{tab:history-inputs}
  \begin{tabular}{@{}p{0.43\linewidth}p{0.53\linewidth}@{}}
\toprule
Models & Observed history used \\
\midrule
MGN; TARNet, CFR \citep{shalit2017}; OLS1, OLS2 \citep{huang2024} & Most recent 16 days of tumor volumes and treatments. Shorter histories are padded by repeating their earliest observations. \\
\addlinespace
RMSN \citep{lim2018}; G-Net \citep{li2021}; CRN, CRN-ERM \citep{bica2020,huang2024}; CT, CT-ERM, EDCT \citep{melnychuk2022,huang2024} & Full available history of tumor volumes and treatments. \\
\addlinespace
MSM \citep{robins2000,lim2018} & Two most recent tumor-volume observations and cumulative treatment counts over the available history. \\
\addlinespace
INSITE \citep{kacprzyk2024} & Available tumor-volume and treatment history for fitting patient-specific equation coefficients. \\
\addlinespace
Persistence & Last observed tumor volume. \\
\addlinespace
Training mean & No patient history at prediction time; returns the training-set mean volume. \\
\bottomrule
\end{tabular}

\end{table}

Our comparison therefore evaluates these model implementations with their respective
history inputs; it does not isolate architecture from history length.

\paragraph{Models not evaluated.} We did not evaluate Causal CPC
\citep{elbouchattaoui2024}, CDM \citep{alinezhad2026}, the G-Transformer
of \citet{xiong2024}, or the causal state-space model of
\citet{abidakun2026}.

\paragraph{Protocol mismatch.} The released configurations of the
G-Transformer of \citet{hess2024} and CausalLongPFN \citep{zare2026}
do not match our evaluation protocol.

\paragraph{Execution errors.} We could not run COSTAR \citep{meng2023}
because its released code produced errors.

\paragraph{The closest prior work.}
\citet{huang2024} ask whether representation balancing improves prediction
accuracy. They report that removing the balancing module performs at least
as well across their study. Our five-day results partly agree
(App.~\ref{app:rosters}): removing representation balancing lowers mean
nRMSE for CRN with both plan constructions at both assignment strengths,
and for CT only at $\gamma = 10$. Their study examines whether a model
component improves prediction performance. We examine what that performance
means: whether low nRMSE indicates that the model has learned the effect
of treatment.

\clearpage
\section{Training and optimization}
\label{app:training}

Training, validation, and test data comprise disjoint cohorts of
$10\,000$, $1\,000$, and $1\,000$ simulated patients, respectively.

For MSM, RMSN, G-Net, CRN, CT and EDCT, we use the Causal Transformer code and
released tumor-simulator hyperparameters \citep{melnychuk2022}. We reuse the
$\gamma = 4$ presets at both $\gamma = 4$ and $\gamma = 10$ without retuning.
Neural training stages use 100 epochs, reducing the released CT and CRN budgets
from 150. Batch sizes and learning rates follow the presets. CT-ERM and CRN-ERM
additionally disable balancing. INSITE uses its authors' code and released
Cancer PKPD constants \citep{kacprzyk2024} across both assignment strengths and
plan constructions without retuning. We could not establish the exact
experimental settings used to tune these constants.

\citet{shalit2017} evaluated TARNet, CFR, OLS1 and OLS2 on static treatment tasks.
\citet{huang2024} subsequently evaluated them on the tumor simulator, but did
not report the TARNet and CFR architectures and training hyperparameters.
We use our own longitudinal adaptations with the settings specified here.
TARNet and CFR use two
shared layers of width 128 with exponential linear unit (ELU) activations and
four treatment-specific heads of hidden width 64. We train them for 10 epochs
with Adam, learning rate 0.001 and batch size 512, retaining final weights.
CFR uses a fixed balancing coefficient of 1 for a group-to-pooled
mean-discrepancy penalty and does not normalize learned representations.
OLS1 and OLS2 use unregularized
least squares separately for each forecast day, conditioning on observed
history and that day's proposed treatment.

We train MGN and its ablations for 10 epochs over the full training
corpus using Adam with a learning rate of 0.001 and a batch size of 32.
The objective is mean squared error on globally standardized tumor
volumes, averaged equally over the five forecast outputs without
reweighting by tumor size. We perform no hyperparameter optimization.

Each variant is fitted once per run and used to predict tumor volumes
under all proposed test plans, with no separate fits for subsets of
plans. Main-text MGN results use the checkpoint from the final epoch,
whereas appendix results use the checkpoint with the lowest validation MSE.

\section{MGN in detail}
\label{app:mgn}

\begin{figure}[H]
  \centering
  \resizebox{\textwidth}{!}{%
  \begin{tikzpicture}[
    font=\small,
    inp/.style={draw=muted, fill=soft, rounded corners=1pt, inner sep=4pt, align=center},
    proc/.style={draw=muted, fill=white, rounded corners=1pt, inner sep=4pt, align=center},
    ar/.style={-{Latex[length=1.9mm]}, draw=muted},
    lbl/.style={font=\scriptsize, color=muted}
  ]
  \node[inp, text width=2.7cm] (vol) at (0,0.75)
    {past volumes\\[1pt] $V_{t-15:t}$};
  \node[inp, text width=2.7cm] (trt) at (0,-0.75)
    {past treatments\\[1pt] $a_{t-15:t}$};
  \node[inp, text width=2.7cm, draw=radio, fill=radio!8] (plan) at (0,-2.0)
    {proposed plan\\[1pt] $a_{t+1:t+\tau}$};

  \node[proc, text width=1.4cm] (revin) at (3.0,0.75) {RevIN\\[1pt] \scriptsize $\mu,\sigma$};
  \node[proc, text width=1.7cm] (gru) at (5.2,0) {\textbf{GRU}\\[1pt] \scriptsize 1 layer, 128};
  \draw[ar] (vol) -- (revin);
  \draw[ar] (revin.east) -- ++(0.35,0) |- (gru.west);
  \draw[ar] (trt.east) -- ++(1.9,0) |- (gru.west);

  \node[proc, minimum width=3.5cm, minimum height=3.5cm] (head) at (9.0,-0.55) {};
  \node[anchor=north, align=center] at ($(head.north)+(0,-0.12)$)
    {\textbf{masked MLP head}\\[1pt] \scriptsize shared weights,\\[-1pt] \scriptsize one call per day $k$};
  \draw[ar] (gru.east) -- node[lbl, above] {$h$} ++(0.6,0) |- ([yshift=0.75cm]head.west);
  \draw[ar] (plan.east) -- ++(0.9,0) |- ([yshift=-0.92cm]head.west);

  \begin{scope}[shift={(head.center)}, xshift=-0.75cm, yshift=-0.38cm]
    \foreach \coli in {1,...,5}{
      \node[lbl] at ({(\coli-1)*0.36}, 0.62) {$a_{t+\coli}$};}
    \foreach \rowi in {1,...,5}{
      \node[lbl, anchor=east] at (-0.28, {0.36-(\rowi-1)*0.36}) {$k{=}\rowi$};
      \foreach \coli in {1,...,5}{
        \ifnum\coli>\rowi\relax
          \draw[muted!45] ({(\coli-1)*0.36-0.12}, {0.48-(\rowi-1)*0.36}) rectangle ++(0.24,-0.24);
        \else
          \fill[radio] ({(\coli-1)*0.36-0.12}, {0.48-(\rowi-1)*0.36}) rectangle ++(0.24,-0.24);
        \fi}
      \draw[ar] ({4*0.36+0.18}, {0.36-(\rowi-1)*0.36}) -- ++(0.42,0)
        node[lbl, right, inner sep=1pt] {$f_{\rowi}$};}
  \end{scope}

  \node[proc, text width=3.7cm, draw=radio] (comp) at (13.2,-0.55)
    {\textbf{multiplicative output}\\[3pt]
     $\hat V_{t+k} = V_t \cdot \exp(f_k)$};
  \draw[ar] (head.east) -- (comp.west);
  \node[lbl, align=center, right=0.6cm of comp] (out) {$\hat V_{t+1}, \dots, \hat V_{t+5}$};
  \draw[ar] (comp.east) -- (out.west);

  \draw[ar, draw=radio] (vol.north) -- ++(0,0.6) -| node[lbl, pos=0.25, above, color=radio]
    {$V_t$: one last observed volume in raw units for all five days}
    (comp.north);
  \end{tikzpicture}}
  \caption{\textbf{MGN architecture.} The encoder takes the past 16 days of tumor volumes and treatments, and emits a hidden state $h$. The masked MLP head is called once per horizon step $k$, with the proposed plan zeroed after day $k$. Each call emits a scalar $f_k$, which is composed multiplicatively on the last observed volume to produce the predicted volume $\hat V_{t+k}$. RevIN stands for reversible instance normalization \citep{kim2022}.}
  \label{fig:arch}
\end{figure}
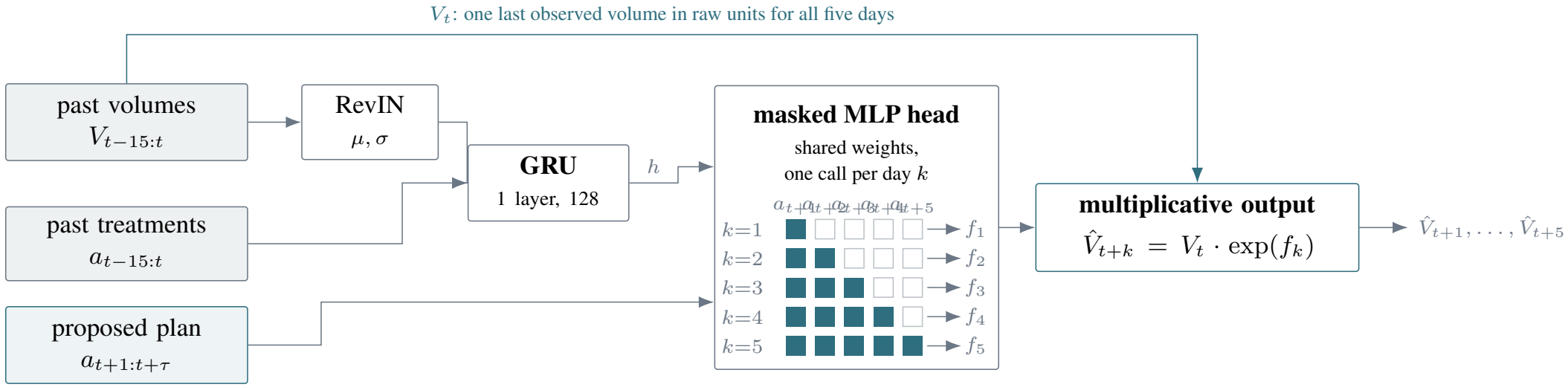

Figure~\ref{fig:arch} summarizes the architecture. The encoder, masked head and multiplicative
output are specified below.

\paragraph{Encoder.} The volume history is normalized by its own window mean and spread,
$z_s = (V_s - \mu) / \sigma$ for $s = t-15, \dots, t$, where $\mu$ and $\sigma$ are the mean and
standard deviation of $V_{t-15:t}$ \citep{kim2022}. It is concatenated day by day with the treatment
indicator, and encoded by a one-layer GRU \citep{cho2014} with $128$ units. We keep its final hidden
state $h$. The normalization acts on the encoder's input only; unlike standard practice it is
never undone on the output, because the output is anchored on the last observed volume instead
(Eq.~\ref{eq:logcompose}).

\paragraph{Masked head.} One two-layer multi-layer perceptron (MLP) \citep{rumelhart1986}
($512$ hidden units, rectified linear unit (ReLU) \citep{nair2010}) is applied once per horizon
step, sharing weights across the five calls. Its input for step $k$ is $h$, a one-hot tag naming
$k$, and the proposed plan with every day after $k$ set to zero. It makes $\partial f_k / \partial a_{t+m}$ exactly zero for $m > k$.

During training, treatment assignment depends on the patient's recent tumor history. The proposed
test plans are chosen independently of that history. The mask tests whether preventing later treatments from entering
earlier predictions improves generalization to those plans. App.~\ref{app:rosters} compares the
masked and unmasked variants.

\paragraph{Multiplicative output.} The head emits a scalar $f_k$ for each forecast day $k$.
Each $f_k$ is added to the log of the last observed volume,
\begin{equation}
  \log \hat V_{t+k} \;=\; \log V_t \;+\; f_k(h,\, a_{t+1:t+k}),
  \qquad\text{i.e.}\qquad
  \hat V_{t+k} \;=\; V_t \cdot \exp\!\big( f_k(h,\, a_{t+1:t+k}) \big).
  \label{eq:logcompose}
\end{equation}
$V_t$ is the last volume observed in the window, in raw units. It has no index $k$: all five
predictions are multiplicative departures from the same origin. Equation~\ref{eq:logcompose}
shares the multiplicative structure of the simulator's evolution equation
(Eq.~\ref{eq:simulator-evolution}).

\paragraph{Static covariates.} The simulator supplies patient type as a static covariate.
Patient type shifts sensitivity to chemotherapy and radiotherapy. MSM, RMSN, G-Net, CRN,
CRN-ERM, CT, CT-ERM, EDCT and INSITE receive it. TARNet, CFR, OLS1, OLS2 and MGN do not.
We report MGN without patient type. In a separate ten-seed ablation, supplying it at every
encoder step, with or without a direct path to the prediction head, did not consistently reduce
aggregate nRMSE under either plan construction at either assignment strength.

\paragraph{Across-seed aggregation.}
We identify each run by a base seed, which the simulator uses directly
to generate the patient cohorts.
Aggregate nRMSE and MAE results are reported as means
and sample standard deviations over ten base seeds.
The plan-variance sweep and all three diagnostic summaries use the same seeds.
The contrast-error and Jacobian figures show equally weighted means over these ten seeds.
For their compact tables, we first average the decile-level statistics within each of the
nonoverlapping ranges 0--5, 6--8 and 9 for each seed, then report the equally weighted
mean and sample standard deviation across seeds. The error-concentration table and
paired comparisons retain two-sided 95\% Student-$t$ intervals with nine degrees of freedom.
Tumor-volume deciles are ranked separately within each seed;
they are not fixed physical-volume bands across seeds. For comparisons with the
plan-independent reference, we first subtract that seed's reference in the same decile,
then summarize these paired differences across seeds.

\section{Full prediction-error results}
\label{app:rosters}

Tables~\ref{tab:wmhs-g10-full} and \ref{tab:wmhs-g4} report results for all
evaluated models. The former restores the five MGN ablations omitted from
Table~\ref{tab:wmhs-g10}; the latter reports results at $\gamma = 4$, within the
ranges evaluated by \citet{melnychuk2022} and in the multi-step experiments of
\citet{bica2020}.

\paragraph{MGN ablations.}
The ablations cross temporal masking with three output parameterizations:
multiplicative composition (Eq.~\ref{eq:logcompose}), additive output in the
window-specific RevIN scale, and additive output in the global volume
scale. All variants use the same training settings (App.~\ref{app:training}). The
masked variants use a shared head evaluated separately at each forecast
step, whereas the unmasked variants emit the full horizon jointly; the
masked--unmasked comparison therefore includes both changes.

\begin{table}[H]\centering\scriptsize
\setlength{\tabcolsep}{3.5pt}
\caption{Complete prediction-error results at $\gamma = 10$ and $\tau = 5$,
including the five MGN ablations omitted from Table~\ref{tab:wmhs-g10}.
Models are sorted by mean nRMSE within each plan construction. Entries are mean $\pm$
sample standard deviation over ten seeds; lower is better, and the best
mean in each column is bold.}
\label{tab:wmhs-g10-full}
\begin{tabular}{l rr @{\qquad} l rr}
\toprule
\multicolumn{3}{c}{\textbf{sliding treatment plans}} & \multicolumn{3}{c}{\textbf{random treatment plans}} \\
\cmidrule(lr){1-3}\cmidrule(lr){4-6}
model & nRMSE \% & MAE & model & nRMSE \% & MAE \\
\midrule
\ourm MGN & \best{3.170}\sd{0.84} & \best{10.357}\sd{4.4}  & \ourm MGN & \best{2.494}\sd{0.60} & \best{8.577}\sd{3.2}  \\
\ourm MGN (unmasked, multiplicative) & 3.262\sd{0.79} & 10.831\sd{4.3}  & \ourm MGN (unmasked, multiplicative) & 2.576\sd{0.63} & 8.778\sd{3.3}  \\
\theirm TARNet & 3.497\sd{0.85} & 12.059\sd{4.2}  & \ourm MGN (masked, additive, global) & 2.748\sd{0.67} & 10.219\sd{3.2}  \\
\ourm MGN (masked, additive, revin) & 3.608\sd{0.94} & 12.061\sd{4.9}  & \theirm TARNet & 2.788\sd{0.65} & 11.490\sd{3.3}  \\
\ourm MGN (unmasked, additive, revin) & 3.639\sd{0.97} & 12.184\sd{5.0}  & \theirm CFR & 2.934\sd{0.82} & 11.802\sd{3.3}  \\
\theirm CFR & 3.853\sd{1.1} & 14.286\sd{5.5}  & \theirm CRN-ERM & 3.072\sd{1.00} & 11.909\sd{7.3}  \\
\ourm MGN (masked, additive, global) & 4.854\sd{1.3} & 16.624\sd{6.6}  & \theirm INSITE & 3.077\sd{0.68} & 15.236\sd{3.2}  \\
\theirm INSITE & 5.004\sd{1.2} & 16.964\sd{5.1}  & \theirm CT-ERM & 3.153\sd{0.64} & 14.442\sd{2.2}  \\
\nullm persistence & 5.128\sd{1.5} & 17.295\sd{7.3}  & \theirm CT & 3.268\sd{1.0} & 15.974\sd{12}  \\
\theirm CT-ERM & 6.112\sd{1.8} & 30.051\sd{5.7}  & \theirm G-Net & 3.296\sd{0.69} & 12.284\sd{3.2}  \\
\theirm MSM & 6.736\sd{1.9} & 25.497\sd{11}  & \ourm MGN (unmasked, additive, global) & 3.403\sd{0.91} & 11.362\sd{3.9}  \\
\theirm CT & 7.146\sd{2.5} & 33.298\sd{21}  & \theirm OLS1 & 3.564\sd{0.81} & 15.398\sd{3.4}  \\
\theirm G-Net & 7.518\sd{2.6} & 29.008\sd{12}  & \theirm OLS2 & 3.673\sd{0.85} & 17.716\sd{3.7}  \\
\theirm CRN-ERM & 7.984\sd{2.6} & 31.800\sd{19}  & \theirm CRN & 3.823\sd{0.98} & 13.583\sd{4.3}  \\
\theirm OLS2 & 8.060\sd{2.1} & 35.121\sd{12}  & \theirm EDCT & 3.926\sd{0.88} & 13.515\sd{4.2}  \\
\ourm MGN (unmasked, additive, global) & 8.098\sd{2.5} & 26.879\sd{12}  & \nullm train mean & 4.640\sd{1.1} & 21.972\sd{4.4}  \\
\theirm CRN & 10.599\sd{3.0} & 44.645\sd{19}  & \ourm MGN (unmasked, additive, revin) & 7.944\sd{2.2} & 28.538\sd{12}  \\
\theirm EDCT & 10.714\sd{2.7} & 42.419\sd{15}  & \ourm MGN (masked, additive, revin) & 8.319\sd{2.4} & 29.179\sd{12}  \\
\theirm OLS1 & 10.761\sd{2.8} & 49.012\sd{17}  & \nullm persistence & 12.713\sd{3.5} & 55.454\sd{22}  \\
\nullm train mean & 12.539\sd{3.3} & 58.851\sd{20}  & \theirm MSM & 14.493\sd{4.0} & 63.741\sd{25}  \\
\theirm RMSN & 20.105\sd{9.9} & 87.664\sd{60}  & \theirm RMSN & 25.420\sd{13} & 102.924\sd{76}  \\
\bottomrule
\end{tabular}
\legend
\end{table}

\paragraph{Results at \texorpdfstring{$\gamma = 4$}{gamma = 4}.}
By mean nRMSE at $\gamma = 4$, persistence
outperforms seven of the thirteen published models with sliding treatment plans
but none with random treatment plans. With random treatment plans, the masked
additive-global ablation has the lowest nRMSE, while MGN has the lowest MAE.

\begin{table}[H]\centering\scriptsize
\setlength{\tabcolsep}{3.5pt}
\caption{Complete prediction-error results at $\gamma = 4$ and $\tau = 5$.
Models are sorted by mean nRMSE within each plan construction. Entries are mean $\pm$
sample standard deviation over ten seeds; lower is better, and the best
mean in each column is bold. Model references are given in
App.~\ref{app:models}.}
\label{tab:wmhs-g4}
\begin{tabular}{l rr @{\qquad} l rr}
\toprule
\multicolumn{3}{c}{\textbf{sliding treatment plans}} & \multicolumn{3}{c}{\textbf{random treatment plans}} \\
\cmidrule(lr){1-3}\cmidrule(lr){4-6}
model & nRMSE \% & MAE & model & nRMSE \% & MAE \\
\midrule
\ourm MGN & \best{1.159}\sd{0.36} & \best{2.289}\sd{1.0}  & \ourm MGN (masked, additive, global) & \best{0.883}\sd{0.16} & 2.519\sd{0.54}  \\
\ourm MGN (unmasked, multiplicative) & 1.208\sd{0.24} & 2.504\sd{0.80}  & \theirm CRN-ERM & 0.936\sd{0.20} & 2.375\sd{0.65}  \\
\ourm MGN (masked, additive, global) & 1.223\sd{0.25} & 2.704\sd{0.81}  & \ourm MGN & 0.940\sd{0.20} & \best{2.031}\sd{0.69}  \\
\theirm CFR & 1.446\sd{0.44} & 3.881\sd{1.4}  & \ourm MGN (unmasked, additive, global) & 0.980\sd{0.19} & 2.634\sd{0.64}  \\
\ourm MGN (masked, additive, revin) & 1.468\sd{0.32} & 2.849\sd{0.91}  & \ourm MGN (unmasked, multiplicative) & 0.983\sd{0.18} & 2.137\sd{0.67}  \\
\ourm MGN (unmasked, additive, revin) & 1.530\sd{0.33} & 3.009\sd{0.96}  & \theirm EDCT & 0.986\sd{0.20} & 2.813\sd{1.1}  \\
\theirm CRN-ERM & 1.570\sd{0.36} & 3.702\sd{1.2}  & \theirm RMSN & 1.051\sd{0.22} & 2.272\sd{0.70}  \\
\theirm TARNet & 1.614\sd{0.35} & 3.770\sd{1.2}  & \theirm TARNet & 1.062\sd{0.23} & 3.208\sd{0.85}  \\
\theirm CT & 1.647\sd{0.42} & 4.647\sd{1.4}  & \theirm CRN & 1.075\sd{0.31} & 3.173\sd{1.7}  \\
\theirm INSITE & 1.649\sd{0.32} & 4.661\sd{0.98}  & \theirm CFR & 1.084\sd{0.23} & 3.587\sd{0.81}  \\
\theirm CT-ERM & 1.701\sd{0.52} & 5.301\sd{2.7}  & \theirm CT & 1.118\sd{0.28} & 3.007\sd{0.91}  \\
\nullm persistence & 1.800\sd{0.45} & 3.572\sd{1.3}  & \theirm INSITE & 1.136\sd{0.22} & 4.700\sd{0.68}  \\
\theirm RMSN & 1.841\sd{0.53} & 3.731\sd{1.6}  & \theirm G-Net & 1.155\sd{0.16} & 3.780\sd{0.41}  \\
\theirm G-Net & 1.843\sd{0.25} & 6.455\sd{0.84}  & \theirm MSM & 1.227\sd{0.31} & 2.890\sd{0.98}  \\
\ourm MGN (unmasked, additive, global) & 2.136\sd{0.51} & 4.128\sd{1.4}  & \theirm CT-ERM & 1.229\sd{0.38} & 3.657\sd{2.0}  \\
\theirm CRN & 2.170\sd{1.1} & 5.971\sd{3.4}  & \theirm OLS2 & 1.395\sd{0.31} & 3.830\sd{0.89}  \\
\theirm MSM & 2.985\sd{0.73} & 8.130\sd{3.0}  & \theirm OLS1 & 1.434\sd{0.32} & 3.684\sd{0.97}  \\
\theirm EDCT & 3.493\sd{4.4} & 23.347\sd{56}  & \nullm train mean & 1.950\sd{0.41} & 9.979\sd{1.1}  \\
\theirm OLS2 & 3.838\sd{0.95} & 10.673\sd{3.8}  & \ourm MGN (masked, additive, revin) & 3.186\sd{0.74} & 6.686\sd{2.5}  \\
\theirm OLS1 & 4.045\sd{1.00} & 11.320\sd{4.0}  & \ourm MGN (unmasked, additive, revin) & 3.265\sd{0.84} & 6.931\sd{2.6}  \\
\nullm train mean & 4.683\sd{1.1} & 17.145\sd{4.0}  & \nullm persistence & 4.588\sd{1.2} & 11.855\sd{4.4}  \\
\bottomrule
\end{tabular}
\legend
\end{table}

\section{Error concentration across models}
\label{app:error-concentration}

Table~\ref{tab:errconc} reports each model's sum of squared errors in the largest
tumor-volume decile divided by its sum over all deciles.
These results summarize all ten base seeds.
The diagnostic uses test windows with all ten proposed plans and finite final-step outcomes
under every plan, as in the contrast analysis.

Across MGN and the thirteen published models, the mean largest-decile share ranges from
$84.4\,\%$ to $94.4\,\%$.

\begin{table}[H]\centering\small
\caption{Largest-decile share of squared error over ten base seeds, with random treatment
plans, $\gamma = 10$ and $\tau = 5$. Entries report the equally weighted mean share and
its 95\% Student-$t$ confidence interval. Deciles are defined by last observed tumor volume
separately within each seed. Model references are given in App.~\ref{app:models}.}
\label{tab:errconc}

\begin{tabular}{lrr}
\toprule
Model & Mean (\%) & 95\% CI \\
\midrule
MGN & 92.8 & [89.4, 96.1] \\
CT & 87.3 & [77.1, 97.5] \\
CT-ERM & 84.4 & [78.8, 89.9] \\
CRN & 94.4 & [92.0, 96.8] \\
CRN-ERM & 92.8 & [88.3, 97.3] \\
EDCT & 94.3 & [90.9, 97.6] \\
G-Net & 90.7 & [88.2, 93.2] \\
MSM & 93.4 & [89.9, 96.9] \\
RMSN & 84.6 & [70.7, 98.5] \\
INSITE & 86.3 & [84.0, 88.6] \\
TARNet & 90.3 & [87.6, 93.0] \\
CFR & 92.2 & [89.9, 94.4] \\
OLS1 & 90.4 & [88.1, 92.7] \\
OLS2 & 89.0 & [87.2, 90.9] \\
Persistence & 93.1 & [89.5, 96.7] \\
Training mean & 92.4 & [89.7, 95.1] \\
\bottomrule
\end{tabular}

\end{table}

\section{How the squared error splits into level and contrast}
\label{app:contrast-split}

Predictions can have the correct mean tumor volume across a window's proposed plans yet fail
to reflect how simulated outcomes differ between those plans. The sum of squared prediction
errors across the window's plans splits exactly into a \emph{level term} and a \emph{contrast term}:
\begin{equation}
\label{eq:split}
  \underbrace{\sum_{a \in \mathcal{A}^i_t}
      \big( \hat V^i_{t+\tau}[a] - V^i_{t+\tau}[a] \big)^2}_{\text{squared error}}
  =
  \underbrace{2\tau \big( \bar{\hat V}^i_t - \bar V^i_t \big)^2}
      _{\text{level}}
  +
  \underbrace{\sum_{a \in \mathcal{A}^i_t}
      \big( \hat V^i_{t+\tau}[a] - \bar{\hat V}^i_t
          - V^i_{t+\tau}[a] + \bar V^i_t \big)^2}
      _{\text{contrast}}
\end{equation}
with $\bar{\hat V}^i_t$ and $\bar V^i_t$ as in \S\ref{sec:contrast-error}. The level term compares
mean predicted and mean simulated tumor volumes across the plans proposed for each patient.
The contrast term scores the
error on the part of the outcome the plan controls, and it is the term
Eq.~\ref{eq:contrast-error} takes the root mean of.

The split is exact because the cross term vanishes. Write the error on one plan as the sum of the
two parts,
\[
  \hat V^i_{t+\tau}[a] - V^i_{t+\tau}[a]
  \;=\; \big( \bar{\hat V}^i_t - \bar V^i_t \big)
  \;+\; \Big( \big( \hat V^i_{t+\tau}[a] - \bar{\hat V}^i_t \big)
      - \big( V^i_{t+\tau}[a] - \bar V^i_t \big) \Big) .
\]
Squaring and summing over the $2\tau$ plans of $\mathcal{A}^i_t$, the cross term is
$2 \big( \bar{\hat V}^i_t - \bar V^i_t \big)$ times the sum of the second part, and each of that
part's two deviations sums to zero over $\mathcal{A}^i_t$ by the definition of the means.

\clearpage
\section{Contrast error across evaluation settings}
\label{app:contrast-cells}

The ten-seed mean contrast-error pattern with random treatment plans is stable across the two
treatment-assignment strengths. MGN remains below the plan-independent reference in every
tumor-volume decile at both $\gamma = 10$ and $\gamma = 4$. With sliding treatment plans,
its contrast error approaches the plan-independent reference.

\begin{figure}[!htbp]
  \centering
  \includegraphics[width=\textwidth]{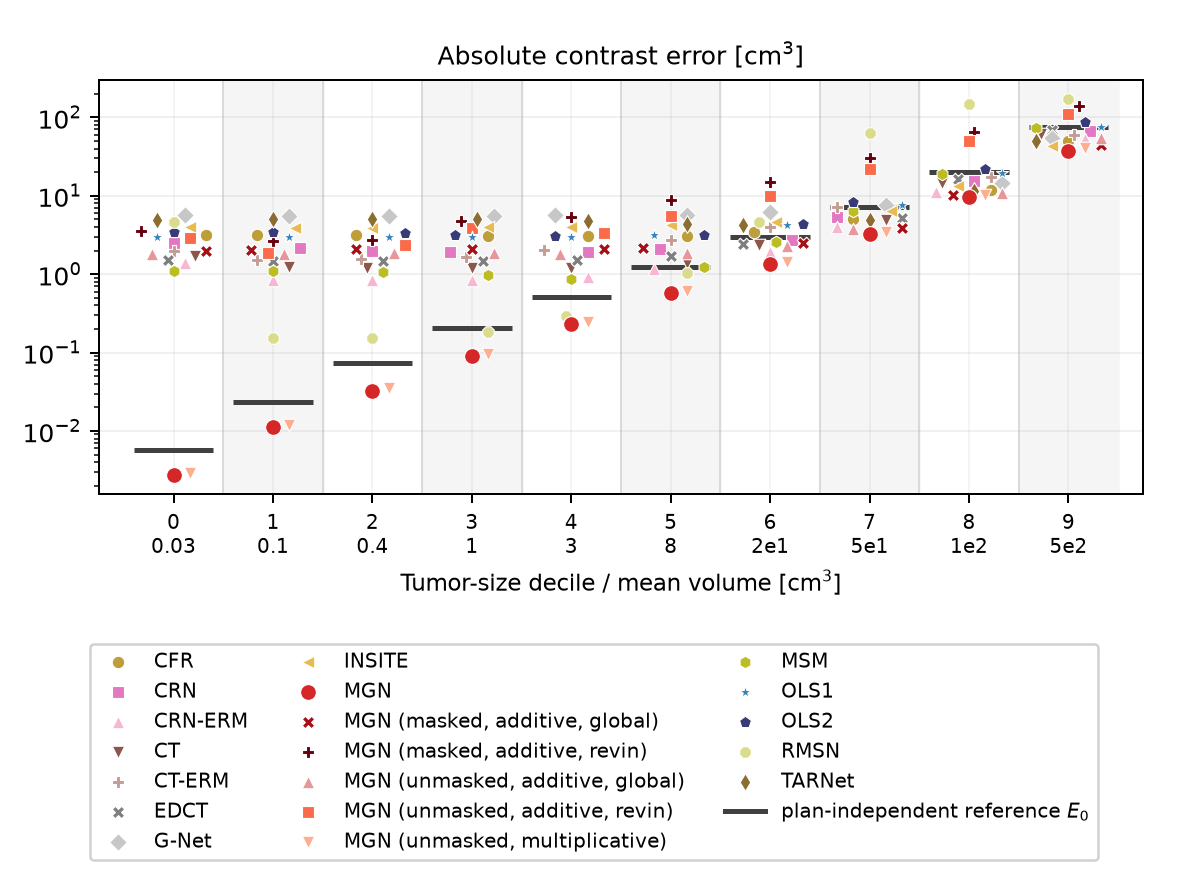}
  \caption[Absolute contrast error: random treatment plans, $\gamma = 10$]{Random treatment
    plans, $\gamma = 10$, $\tau = 5$. Points and reference segments show means over ten base seeds
    of absolute contrast error by tumor-size
    decile. Below each decile index, the second x-axis row gives the mean last observed tumor
    volume in that decile, averaged over ten seeds, in $\mathrm{cm^3}$. Display conventions follow
    Figure~\ref{fig:contrast-by-size}. This figure also
    includes the five MGN ablations.}
  \label{fig:contrast-random-g10-full}
\end{figure}

\begin{figure}[p]
  \centering
  \includegraphics[width=\textwidth]{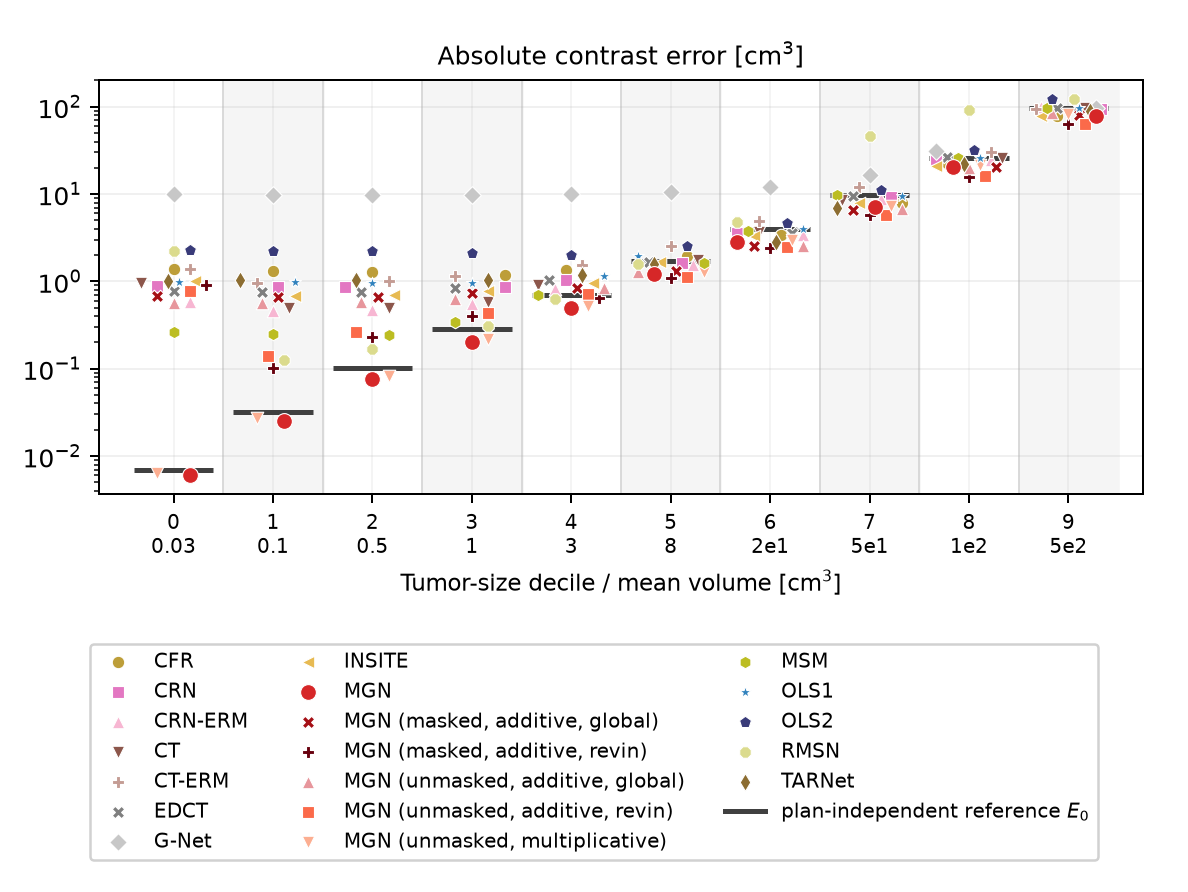}
  \caption[Absolute contrast error: sliding treatment plans, $\gamma = 10$]{Sliding treatment
    plans, $\gamma = 10$, $\tau = 5$. Points and reference segments show means over ten base seeds
    of absolute contrast error by tumor-size
    decile. Below each decile index, the second x-axis row gives the mean last observed tumor
    volume in that decile, averaged over ten seeds, in $\mathrm{cm^3}$. Display conventions follow
    Figure~\ref{fig:contrast-by-size}.}
  \label{fig:contrast-sliding-g10}
\end{figure}

\begin{figure}[p]
  \centering
  \includegraphics[width=\textwidth]{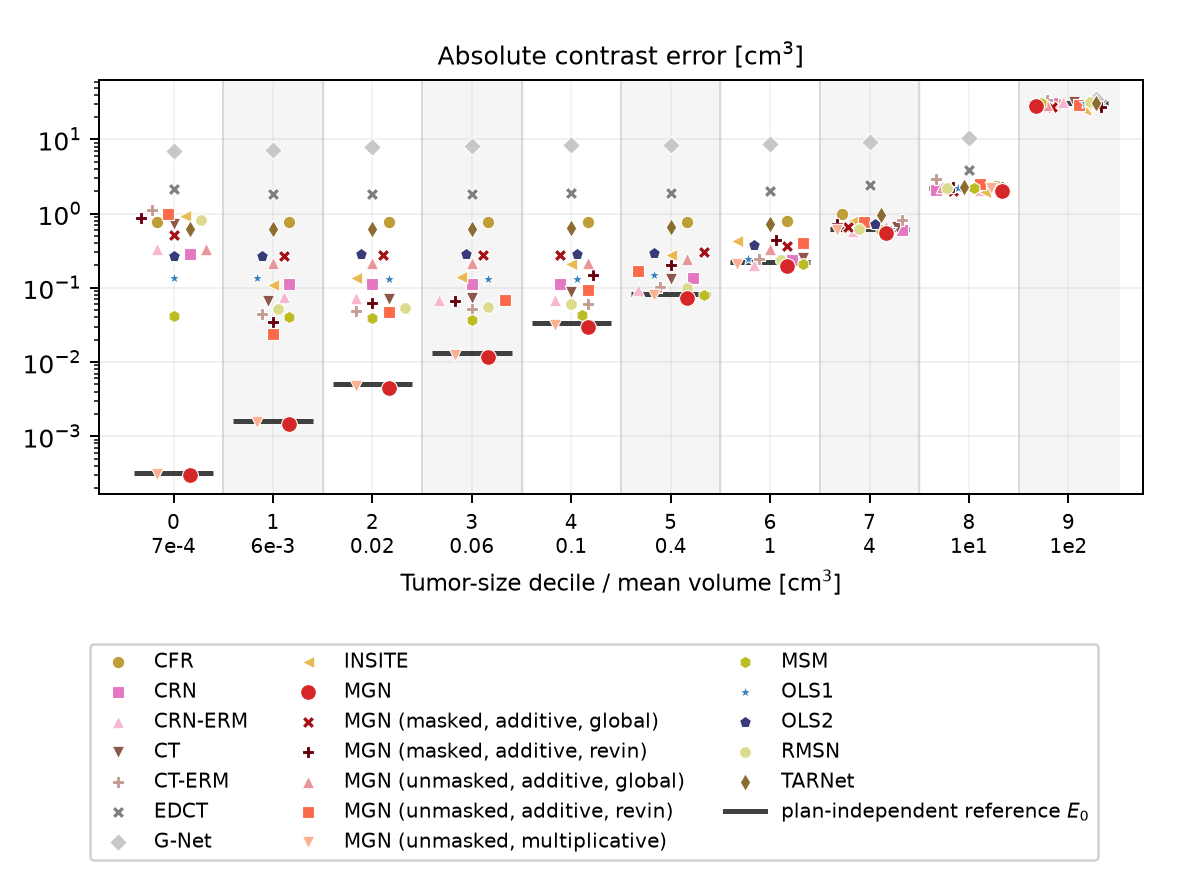}
  \caption[Absolute contrast error: sliding treatment plans, $\gamma = 4$]{Sliding treatment
    plans, $\gamma = 4$, $\tau = 5$. Points and reference segments show means over ten base seeds
    of absolute contrast error by tumor-size
    decile. Below each decile index, the second x-axis row gives the mean last observed tumor
    volume in that decile, averaged over ten seeds, in $\mathrm{cm^3}$. Display conventions follow
    Figure~\ref{fig:contrast-by-size}.}
  \label{fig:contrast-sliding-g4}
\end{figure}

\begin{figure}[p]
  \centering
  \includegraphics[width=\textwidth]{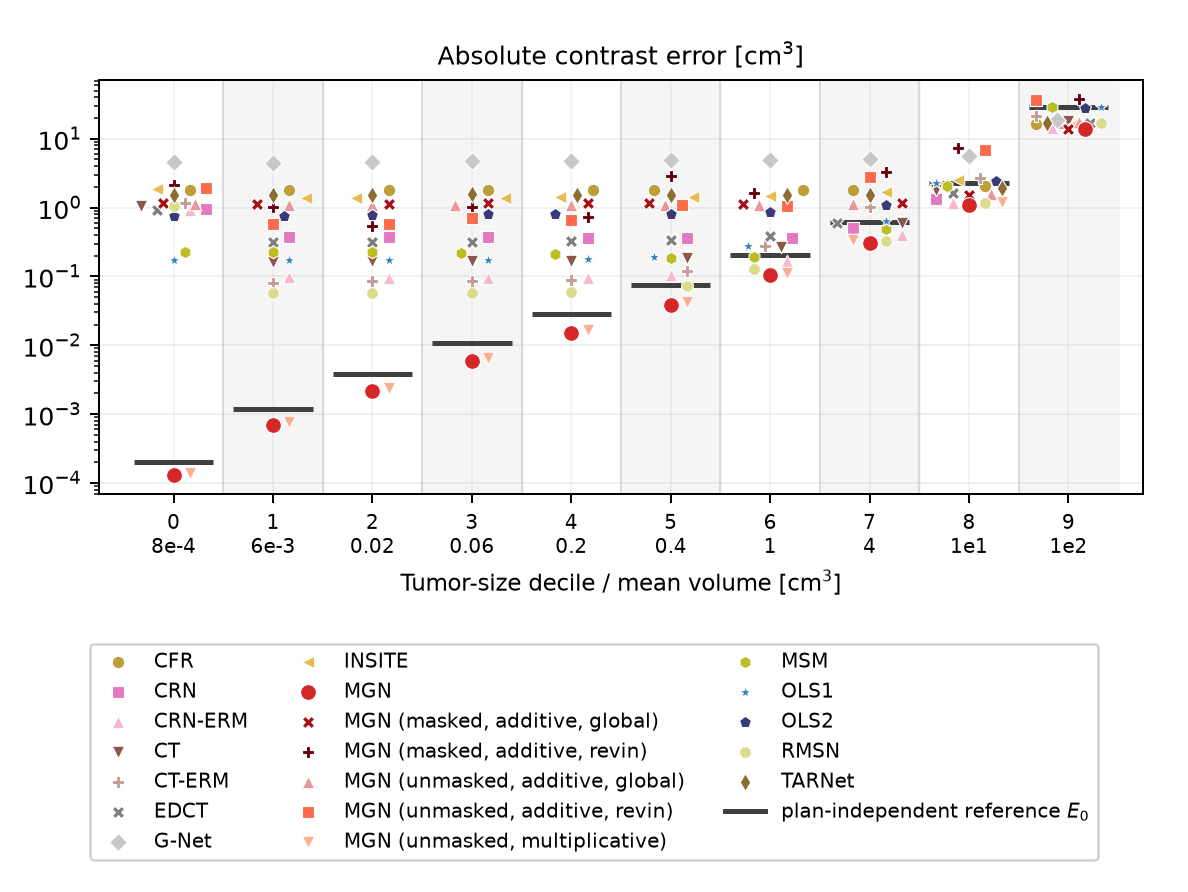}
  \caption[Absolute contrast error: random treatment plans, $\gamma = 4$]{Random treatment
    plans, $\gamma = 4$, $\tau = 5$. Points and reference segments show means over ten base seeds
    of absolute contrast error by tumor-size
    decile. Below each decile index, the second x-axis row gives the mean last observed tumor
    volume in that decile, averaged over ten seeds, in $\mathrm{cm^3}$. Display conventions follow
    Figure~\ref{fig:contrast-by-size}.}
  \label{fig:contrast-random-g4}
\end{figure}

Figures~\ref{fig:contrast-random-g10-full}--\ref{fig:contrast-random-g4} report all four
evaluation settings formed by crossing plan construction and $\gamma$. Each reports absolute contrast error in $\mathrm{cm^3}$ by tumor-volume decile. The first panel uses the same evaluation setting as
Figure~\ref{fig:contrast-by-size} and includes the five MGN ablations from
App.~\ref{app:rosters}. Persistence
and the training mean are omitted because their predictions do not vary across plans, making their
contrast error exactly $E_0$. Axes are scaled separately, so comparisons across panels should use
each model's position relative to $E_0$ rather than its vertical position.

\clearpage
\paragraph{Across-seed summaries.}
Tables~\ref{tab:seed-contrast-g10-random}--\ref{tab:seed-contrast-g4-random}
report mean contrast errors and sample standard deviations across the ten base seeds,
including the plan-independent reference and the five MGN ablations.
Within each seed, decile-level errors are averaged equally over ranges 0--5, 6--8 and 9.
Cells show the across-seed mean plus or minus one standard deviation.
Comparisons with the reference use paired differences
within each seed, as described in App.~\ref{app:mgn}.
In the main setting, RMSN's mean differences from the reference in deciles 3 and 4 are
$-0.021$ and $-0.221\,\mathrm{cm^3}$, respectively. It beats the reference in five of ten
seeds in decile 3 and six of ten in decile 4. The paired 95\% intervals are
$[-0.126, 0.085]$ and $[-0.443, 0.001]\,\mathrm{cm^3}$, respectively; both include zero.
MGN beats the reference in every decile in all ten main-setting seeds.

\begin{table}[!htbp]\centering\footnotesize
\setlength{\tabcolsep}{3pt}
\caption{Mean decile-level contrast error over ten base seeds: random treatment plans, $\gamma = 10$, $\tau = 5$. Units are $\mathrm{cm^3}$. Within each seed, decile-level statistics are averaged equally within ranges 0--5, 6--8 and 9. Cells give the across-seed mean $\pm$ sample standard deviation. Rows are sorted by increasing mean in deciles 0--5. Bold means mark the lowest contrast error in each range. Model references are in App.~\ref{app:models}.}
\label{tab:seed-contrast-g10-random}
\begin{tabular}{lccc}
\toprule
Model & Deciles 0--5 & Deciles 6--8 & Decile 9 \\
\midrule
MGN & $(\mathbf{0.16}\pm0.14)$ & $(\mathbf{4.8}\pm2.9)$ & $(\mathbf{38}\pm9.6)$ \\[3pt]
MGN (unmasked, multiplicative) & $(0.17\pm0.15)$ & $(5\pm3)$ & $(40\pm9.2)$ \\[3pt]
Reference $E_0$ & $(0.34\pm0.31)$ & $(10\pm6.2)$ & $(75\pm18)$ \\[3pt]
CRN-ERM & $(1\pm1.2)$ & $(5.7\pm4)$ & $(53\pm15)$ \\[3pt]
MSM & $(1.1\pm0.055)$ & $(9.4\pm6)$ & $(74\pm18)$ \\[3pt]
RMSN & $(1.1\pm0.51)$ & $(72\pm77)$ & $(170\pm85)$ \\[3pt]
CT & $(1.3\pm0.69)$ & $(7.3\pm4.7)$ & $(61\pm17)$ \\[3pt]
EDCT & $(1.5\pm1.1)$ & $(8.1\pm4.8)$ & $(69\pm17)$ \\[3pt]
MGN (unmasked, additive, global) & $(1.8\pm0.34)$ & $(5.7\pm2.9)$ & $(55\pm14)$ \\[3pt]
CT-ERM & $(1.9\pm1.5)$ & $(9.6\pm3.1)$ & $(59\pm11)$ \\[3pt]
MGN (masked, additive, global) & $(2.1\pm0.44)$ & $(5.5\pm2.5)$ & $(44\pm11)$ \\[3pt]
CRN & $(2.1\pm1.3)$ & $(7.9\pm4.3)$ & $(68\pm17)$ \\[3pt]
OLS1 & $(3\pm0.16)$ & $(10\pm5.6)$ & $(74\pm18)$ \\[3pt]
CFR & $(3.1\pm0.93)$ & $(6.7\pm2.4)$ & $(50\pm15)$ \\[3pt]
OLS2 & $(3.3\pm0.54)$ & $(12\pm6.2)$ & $(87\pm21)$ \\[3pt]
MGN (unmasked, additive, RevIN) & $(3.3\pm2.2)$ & $(27\pm13)$ & $(110\pm33)$ \\[3pt]
INSITE & $(4\pm0.27)$ & $(8.1\pm3)$ & $(43\pm10)$ \\[3pt]
MGN (masked, additive, RevIN) & $(4.6\pm3.1)$ & $(37\pm22)$ & $(140\pm52)$ \\[3pt]
TARNet & $(4.9\pm2)$ & $(6.9\pm2.4)$ & $(50\pm14)$ \\[3pt]
G-Net & $(5.6\pm0.93)$ & $(9.4\pm2.4)$ & $(54\pm15)$ \\[3pt]
\bottomrule
\end{tabular}

\end{table}

\begin{table}[!htbp]\centering\footnotesize
\setlength{\tabcolsep}{3pt}
\caption{Mean decile-level contrast error over ten base seeds: sliding treatment plans, $\gamma = 10$, $\tau = 5$. Units are $\mathrm{cm^3}$. Within each seed, decile-level statistics are averaged equally within ranges 0--5, 6--8 and 9. Cells give the across-seed mean $\pm$ sample standard deviation. Rows are sorted by increasing mean in deciles 0--5. Bold means mark the lowest contrast error in each range. Model references are in App.~\ref{app:models}.}
\label{tab:seed-contrast-g10-sliding}
\begin{tabular}{lccc}
\toprule
Model & Deciles 0--5 & Deciles 6--8 & Decile 9 \\
\midrule
MGN & $(\mathbf{0.34}\pm0.3)$ & $(10\pm6.3)$ & $(79\pm21)$ \\[3pt]
MGN (unmasked, multiplicative) & $(0.35\pm0.31)$ & $(11\pm6.5)$ & $(83\pm19)$ \\[3pt]
Reference $E_0$ & $(0.47\pm0.43)$ & $(13\pm8.6)$ & $(96\pm26)$ \\[3pt]
MGN (masked, additive, RevIN) & $(0.56\pm0.31)$ & $(\mathbf{7.9}\pm4.7)$ & $(\mathbf{63}\pm14)$ \\[3pt]
MSM & $(0.57\pm0.34)$ & $(13\pm8.6)$ & $(96\pm26)$ \\[3pt]
MGN (unmasked, additive, RevIN) & $(0.58\pm0.3)$ & $(8.1\pm4.5)$ & $(64\pm14)$ \\[3pt]
CRN-ERM & $(0.74\pm0.72)$ & $(12\pm8.3)$ & $(100\pm28)$ \\[3pt]
MGN (unmasked, additive, global) & $(0.75\pm0.16)$ & $(9.8\pm5.9)$ & $(84\pm21)$ \\[3pt]
MGN (masked, additive, global) & $(0.82\pm0.26)$ & $(9.8\pm6.1)$ & $(81\pm20)$ \\[3pt]
RMSN & $(0.84\pm0.33)$ & $(47\pm44)$ & $(120\pm18)$ \\[3pt]
CT & $(0.87\pm0.35)$ & $(13\pm8.6)$ & $(97\pm25)$ \\[3pt]
INSITE & $(0.96\pm0.29)$ & $(11\pm6.6)$ & $(79\pm20)$ \\[3pt]
EDCT & $(0.97\pm0.43)$ & $(13\pm8.8)$ & $(97\pm27)$ \\[3pt]
CRN & $(1\pm0.62)$ & $(13\pm8.4)$ & $(95\pm27)$ \\[3pt]
TARNet & $(1.1\pm0.35)$ & $(11\pm6.3)$ & $(90\pm24)$ \\[3pt]
OLS1 & $(1.2\pm0.22)$ & $(13\pm8.4)$ & $(96\pm26)$ \\[3pt]
CFR & $(1.4\pm0.55)$ & $(11\pm6.7)$ & $(79\pm17)$ \\[3pt]
CT-ERM & $(1.4\pm0.9)$ & $(16\pm7.5)$ & $(95\pm21)$ \\[3pt]
OLS2 & $(2.2\pm0.14)$ & $(16\pm10)$ & $(120\pm32)$ \\[3pt]
G-Net & $(10\pm1.1)$ & $(20\pm6.2)$ & $(97\pm24)$ \\[3pt]
\bottomrule
\end{tabular}

\end{table}

\begin{table}[!htbp]\centering\footnotesize
\setlength{\tabcolsep}{3pt}
\caption{Mean decile-level contrast error over ten base seeds: sliding treatment plans, $\gamma = 4$, $\tau = 5$. Units are $\mathrm{cm^3}$. Within each seed, decile-level statistics are averaged equally within ranges 0--5, 6--8 and 9. Cells give the across-seed mean $\pm$ sample standard deviation. Rows are sorted by increasing mean in deciles 0--5. Bold means mark the lowest contrast error in each range. Model references are in App.~\ref{app:models}.}
\label{tab:seed-contrast-g4-sliding}
\begin{tabular}{lccc}
\toprule
Model & Deciles 0--5 & Deciles 6--8 & Decile 9 \\
\midrule
MGN & $(\mathbf{0.02}\pm0.016)$ & $(\mathbf{0.92}\pm0.55)$ & $(28\pm7.6)$ \\[3pt]
MGN (unmasked, multiplicative) & $(0.022\pm0.017)$ & $(1\pm0.61)$ & $(29\pm8.3)$ \\[3pt]
Reference $E_0$ & $(0.023\pm0.018)$ & $(1\pm0.64)$ & $(31\pm9.3)$ \\[3pt]
MSM & $(0.047\pm0.0086)$ & $(1\pm0.63)$ & $(31\pm9.3)$ \\[3pt]
CRN-ERM & $(0.12\pm0.028)$ & $(0.96\pm0.6)$ & $(32\pm10)$ \\[3pt]
OLS1 & $(0.14\pm0.01)$ & $(1\pm0.61)$ & $(31\pm9.3)$ \\[3pt]
CRN & $(0.15\pm0.084)$ & $(0.97\pm0.6)$ & $(31\pm9.2)$ \\[3pt]
RMSN & $(0.19\pm0.066)$ & $(1\pm0.63)$ & $(32\pm8.9)$ \\[3pt]
CT & $(0.19\pm0.088)$ & $(1\pm0.58)$ & $(32\pm10)$ \\[3pt]
MGN (masked, additive, RevIN) & $(0.23\pm0.1)$ & $(1.1\pm0.56)$ & $(27\pm7.6)$ \\[3pt]
MGN (unmasked, additive, RevIN) & $(0.23\pm0.07)$ & $(1.2\pm0.63)$ & $(29\pm8)$ \\[3pt]
CT-ERM & $(0.24\pm0.12)$ & $(1.3\pm1)$ & $(34\pm11)$ \\[3pt]
MGN (unmasked, additive, global) & $(0.24\pm0.059)$ & $(1.1\pm0.6)$ & $(29\pm9)$ \\[3pt]
OLS2 & $(0.28\pm0.043)$ & $(1.1\pm0.56)$ & $(31\pm9.4)$ \\[3pt]
INSITE & $(0.3\pm0.087)$ & $(1.1\pm0.46)$ & $(\mathbf{25}\pm6.9)$ \\[3pt]
MGN (masked, additive, global) & $(0.32\pm0.054)$ & $(1\pm0.46)$ & $(27\pm7.7)$ \\[3pt]
TARNet & $(0.63\pm0.26)$ & $(1.3\pm0.48)$ & $(30\pm8.1)$ \\[3pt]
CFR & $(0.77\pm0.24)$ & $(1.4\pm0.52)$ & $(30\pm7.9)$ \\[3pt]
EDCT & $(1.9\pm5.5)$ & $(2.8\pm5.6)$ & $(31\pm9.1)$ \\[3pt]
G-Net & $(7.9\pm1.4)$ & $(9.5\pm1.2)$ & $(35\pm7.6)$ \\[3pt]
\bottomrule
\end{tabular}

\end{table}

\begin{table}[!htbp]\centering\footnotesize
\setlength{\tabcolsep}{3pt}
\caption{Mean decile-level contrast error over ten base seeds: random treatment plans, $\gamma = 4$, $\tau = 5$. Units are $\mathrm{cm^3}$. Within each seed, decile-level statistics are averaged equally within ranges 0--5, 6--8 and 9. Cells give the across-seed mean $\pm$ sample standard deviation. Rows are sorted by increasing mean in deciles 0--5. Bold means mark the lowest contrast error in each range. Model references are in App.~\ref{app:models}.}
\label{tab:seed-contrast-g4-random}
\begin{tabular}{lccc}
\toprule
Model & Deciles 0--5 & Deciles 6--8 & Decile 9 \\
\midrule
MGN & $(\mathbf{0.01}\pm0.0081)$ & $(\mathbf{0.5}\pm0.28)$ & $(14\pm3.2)$ \\[3pt]
MGN (unmasked, multiplicative) & $(0.012\pm0.0093)$ & $(0.55\pm0.31)$ & $(15\pm3.1)$ \\[3pt]
Reference $E_0$ & $(0.02\pm0.016)$ & $(1\pm0.59)$ & $(29\pm7.1)$ \\[3pt]
OLS1 & $(0.18\pm0.029)$ & $(1.1\pm0.57)$ & $(29\pm7.1)$ \\[3pt]
MSM & $(0.22\pm0.013)$ & $(0.92\pm0.53)$ & $(29\pm7.1)$ \\[3pt]
RMSN & $(0.23\pm0.065)$ & $(0.54\pm0.32)$ & $(17\pm4.2)$ \\[3pt]
CRN-ERM & $(0.23\pm0.032)$ & $(0.58\pm0.3)$ & $(14\pm3.3)$ \\[3pt]
CT-ERM & $(0.27\pm0.082)$ & $(1.3\pm1.3)$ & $(21\pm7.6)$ \\[3pt]
CT & $(0.32\pm0.1)$ & $(0.88\pm0.35)$ & $(18\pm5.7)$ \\[3pt]
EDCT & $(0.42\pm0.28)$ & $(0.87\pm0.44)$ & $(17\pm4.4)$ \\[3pt]
CRN & $(0.47\pm0.34)$ & $(0.73\pm0.4)$ & $(17\pm6.5)$ \\[3pt]
OLS2 & $(0.79\pm0.13)$ & $(1.5\pm0.35)$ & $(28\pm6.8)$ \\[3pt]
MGN (unmasked, additive, RevIN) & $(0.93\pm0.64)$ & $(3.5\pm3.4)$ & $(36\pm19)$ \\[3pt]
MGN (unmasked, additive, global) & $(1.1\pm0.19)$ & $(1.3\pm0.21)$ & $(17\pm4.1)$ \\[3pt]
MGN (masked, additive, global) & $(1.2\pm0.24)$ & $(1.3\pm0.25)$ & $(\mathbf{14}\pm3.1)$ \\[3pt]
MGN (masked, additive, RevIN) & $(1.4\pm0.93)$ & $(4.1\pm2.3)$ & $(38\pm14)$ \\[3pt]
INSITE & $(1.5\pm0.071)$ & $(1.9\pm0.25)$ & $(16\pm3.2)$ \\[3pt]
TARNet & $(1.5\pm0.47)$ & $(1.7\pm0.45)$ & $(17\pm3.8)$ \\[3pt]
CFR & $(1.8\pm0.41)$ & $(1.9\pm0.38)$ & $(16\pm3.5)$ \\[3pt]
G-Net & $(4.7\pm1)$ & $(5.3\pm0.92)$ & $(19\pm3.7)$ \\[3pt]
\bottomrule
\end{tabular}

\end{table}

\clearpage
\section{How the action Jacobian is computed}
\label{app:jacobian-method}

We compute gradients of the predicted final-day tumor volume with respect to the future
treatment inputs at the no-treatment plan, holding the observed history fixed. Treatment inputs
are continuously relaxed as in \S\ref{sec:jacobian}. For one-hot inputs, increasing a treatment
input decreases the no-treatment input by the same amount. For the binary inputs used
by RMSN and MSM, we vary each treatment input independently.

We use automatic differentiation for MGN, RMSN, CT, CT-ERM, CRN, CRN-ERM and EDCT in evaluation
mode. Gradients propagate through the full prediction computation, including autoregressive
feedback where applicable. OLS1 and MSM are affine in the treatment inputs, so their gradients
are constant with respect to the proposed plan.
The ten-seed local-response figures and tables use fresh fits under the original diagnostic
configurations, including base seed 10.
For the six published neural models, each training stage retains the original
100-epoch budget.
The contrast summaries reuse saved predictions where available.

The simulator reference is computed analytically from the discrete-time evolution equation of
\citet[Eq.~(6)]{lim2018}, based on the tumor-growth model of \citet{geng2017}:
\begin{equation}
  V_s = V_{s-1}\bigl[1 + G(V_{s-1}) - \beta_c C_s
  - (\alpha d_s + \beta d_s^2) + \varepsilon_s\bigr],
  \label{eq:simulator-evolution}
\end{equation}
where $G$ is the tumor-growth term, $C_s$ is the chemotherapy concentration, $d_s$ is the
radiotherapy dose and $\varepsilon_s$ is simulation noise. The coefficients $\beta_c$, $\alpha$
and $\beta$ describe the patient's treatment response. Differentiating with respect to the dose
variable $u_{r,j}$ for treatment $r$ on future day $j$ at the no-treatment plan gives
\[
  \frac{\partial V_s}{\partial u_{r,j}}
  = \bigl[B_s + V_{s-1} G'(V_{s-1})\bigr]\frac{\partial V_{s-1}}{\partial u_{r,j}}
    - V_{s-1}\left(\beta_c\frac{\partial C_s}{\partial u_{r,j}}
    + \alpha\frac{\partial d_s}{\partial u_{r,j}}\right),
\]
where $B_s = 1 + G(V_{s-1}) - \beta_c C_s + \varepsilon_s$.
We propagate this derivative along each patient's untreated trajectory, with patient parameters
and simulation noise held fixed. The dose inputs enter through
$C_s = 0.5C_{s-1} + 5u_{\mathrm{chemo},s}$ and $d_s = 2u_{\mathrm{radio},s}$.
Model and simulator derivatives use the same dose units and are aggregated as defined in
\S\ref{sec:jacobian}.

\section{Average local dose response in the other evaluation settings}
\label{app:jacobian-cells}

\begin{figure}[p]
  \centering
  \includegraphics[width=\textwidth]{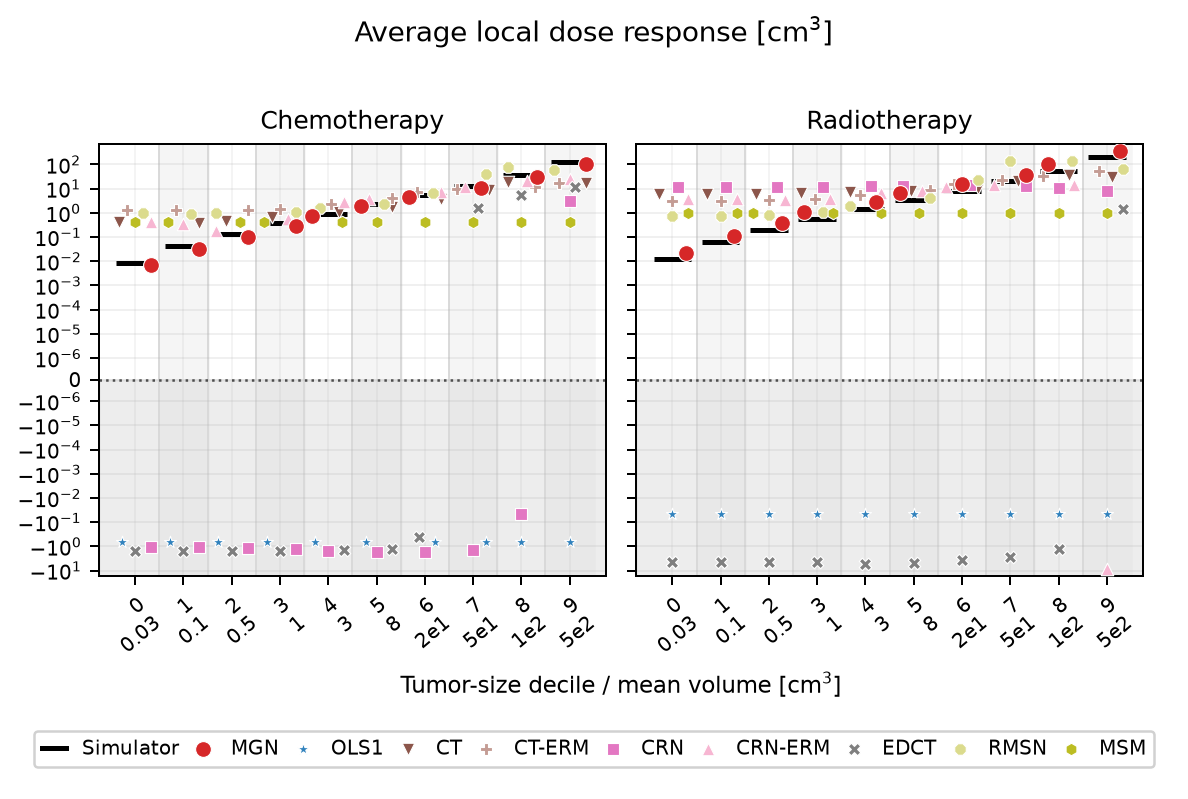}
  \caption[Average local dose response, sliding treatment plans]{Sliding treatment plans,
    $\gamma = 10$, $\tau = 5$. Points and simulator reference segments show means over ten base seeds
    of average local dose response by tumor-size decile.
    Below each decile index, the second x-axis row gives the mean last observed tumor volume in
    that decile, averaged over ten seeds, in $\mathrm{cm^3}$.}
  \label{fig:jacobian-sliding-g10}
\end{figure}

\begin{figure}[p]
  \centering
  \includegraphics[width=\textwidth]{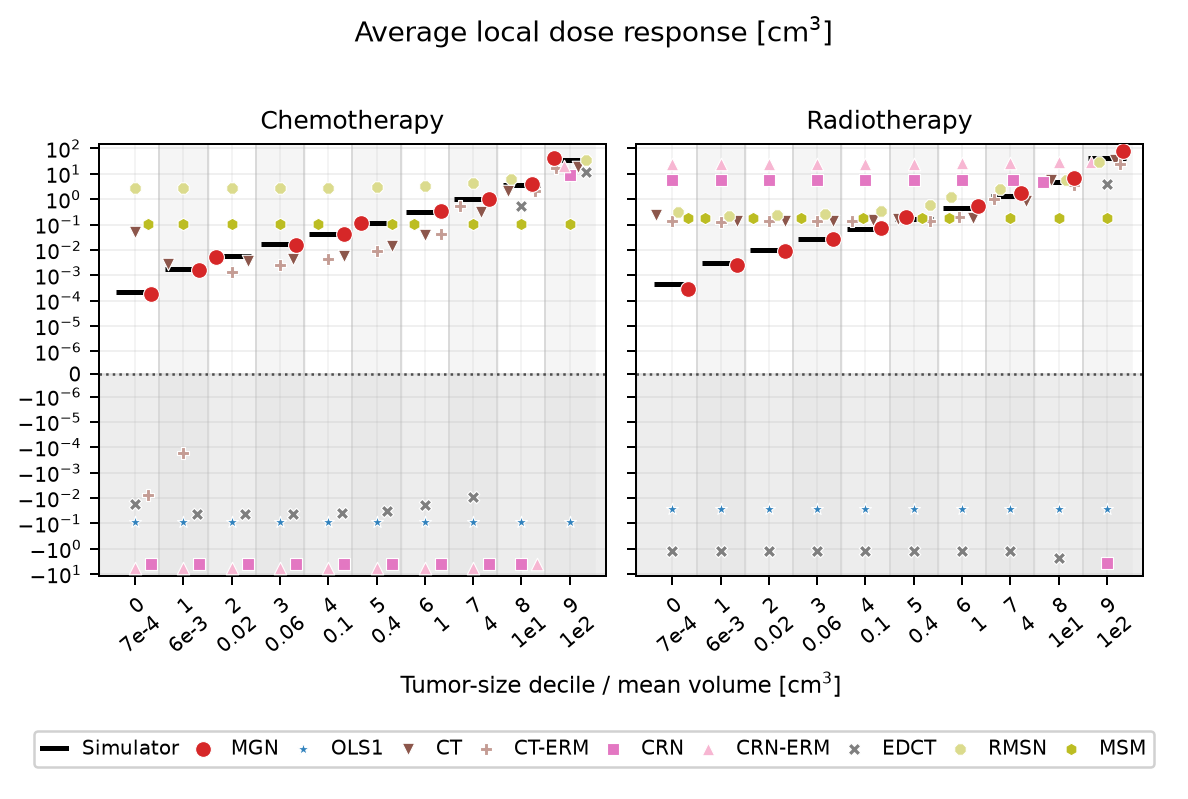}
  \caption[Average local dose response, sliding treatment plans]{Sliding treatment plans,
    $\gamma = 4$, $\tau = 5$. Points and simulator reference segments show means over ten base seeds
    of average local dose response by tumor-size decile.
    Below each decile index, the second x-axis row gives the mean last observed tumor volume in
    that decile, averaged over ten seeds, in $\mathrm{cm^3}$.}
  \label{fig:jacobian-sliding-g4}
\end{figure}

\begin{figure}[p]
  \centering
  \includegraphics[width=\textwidth]{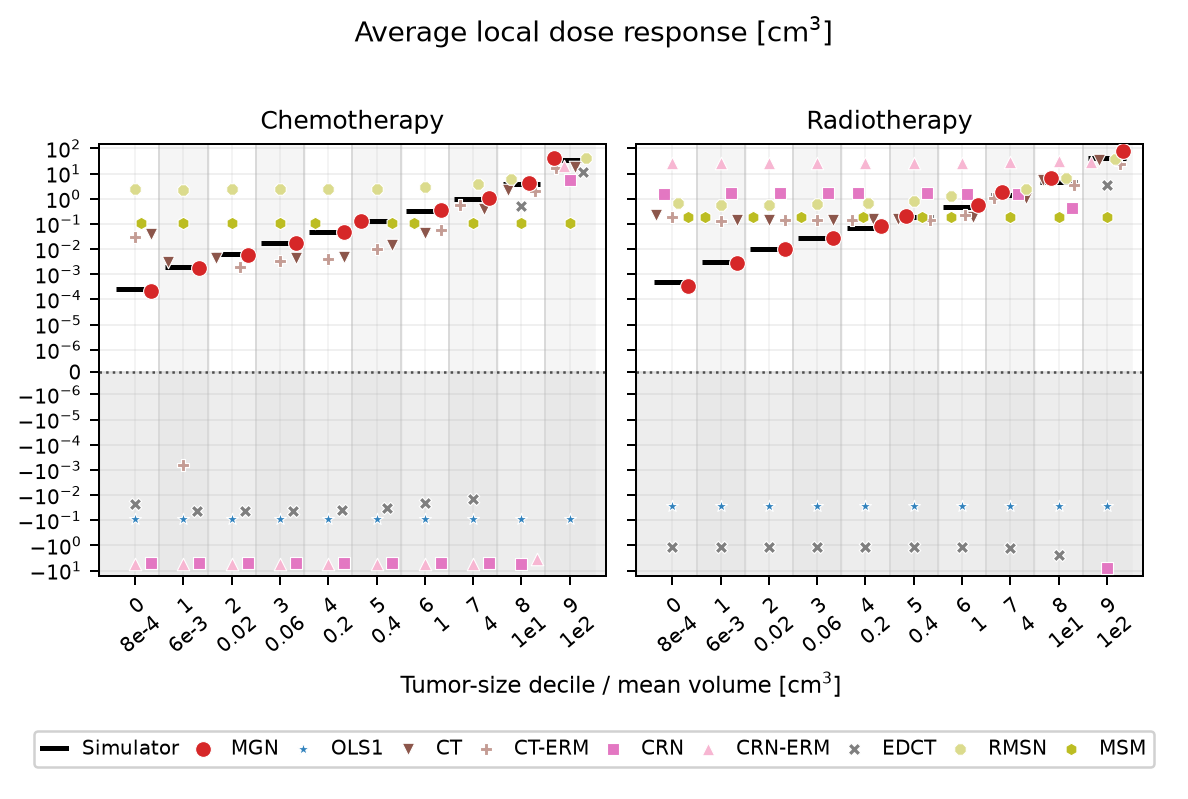}
  \caption[Average local dose response, random treatment plans]{Random treatment plans,
    $\gamma = 4$, $\tau = 5$. Points and simulator reference segments show means over ten base seeds
    of average local dose response by tumor-size decile.
    Below each decile index, the second x-axis row gives the mean last observed tumor volume in
    that decile, averaged over ten seeds, in $\mathrm{cm^3}$.}
  \label{fig:jacobian-random-g4}
\end{figure}

Figures~\ref{fig:jacobian-sliding-g10}--\ref{fig:jacobian-random-g4} repeat
Figure~\ref{fig:jacobian-grid} in the three remaining evaluation settings formed by crossing
plan construction and $\gamma$. Every panel uses the same models,
ten seeds, horizon and scored step. Because every evaluation setting takes the derivative at the
no-treatment plan, the plan construction changes only which test windows enter the calculation.
The display conventions follow Figure~\ref{fig:jacobian-grid}.

\clearpage
\paragraph{Across-seed summaries.}
Tables~\ref{tab:seed-response-g10-random}--\ref{tab:seed-response-g4-random}
report mean local dose responses and sample standard deviations across all ten base seeds
for chemotherapy and radiotherapy in every evaluation setting, including the analytic
simulator reference. Each seed contributes the arithmetic mean of its decile-level responses
within ranges 0--5, 6--8 and 9, with the same windows used for every model and the simulator.
Cells show the across-seed mean plus or minus one standard deviation; positive values
indicate tumor reduction.
MGN's mean radiotherapy responses exceed the simulator's in every main-setting decile.
In the largest decile, the means are $350$ and $186\,\mathrm{cm^3}$, respectively.
With $\gamma = 4$ and random plans, the radiotherapy means also exceed the simulator's
in the larger deciles.

\begin{table}[!htbp]\centering\footnotesize
\setlength{\tabcolsep}{3pt}
\caption{Average local dose response over ten base seeds: random treatment plans, $\gamma = 10$, $\tau = 5$. Units are $\mathrm{cm^3}$. Within each seed, decile-level statistics are averaged equally within ranges 0--5, 6--8 and 9. Cells give the across-seed mean $\pm$ sample standard deviation. Rows are sorted by increasing chemotherapy mean in deciles 0--5. Positive responses indicate tumor reduction. Bold means are closest to the simulator's mean. Model references are in App.~\ref{app:models}.}
\label{tab:seed-response-g10-random}
\begin{tabular}{lcccccc}
\toprule
 & \multicolumn{3}{c}{Chemotherapy} & \multicolumn{3}{c}{Radiotherapy} \\
\cmidrule(lr){2-4}\cmidrule(lr){5-7}
Model & Deciles 0--5 & Deciles 6--8 & Decile 9 & Deciles 0--5 & Deciles 6--8 & Decile 9 \\
\midrule
CRN & $(-2.2\pm7.4)$ & $(-1.2\pm9)$ & $(1.1\pm8)$ & $(-19\pm100)$ & $(-24\pm120)$ & $(-21\pm91)$ \\[3pt]
EDCT & $(-1.5\pm5.2)$ & $(2.3\pm7.1)$ & $(11\pm8.2)$ & $(-4.7\pm13)$ & $(-2.5\pm14)$ & $(1.4\pm13)$ \\[3pt]
OLS1 & $(-0.66\pm0.018)$ & $(-0.66\pm0.018)$ & $(-0.66\pm0.018)$ & $(-0.044\pm0.022)$ & $(-0.044\pm0.022)$ & $(-0.044\pm0.022)$ \\[3pt]
MSM & $(0.43\pm0.014)$ & $(0.43\pm0.014)$ & $(0.43\pm0.014)$ & $(\mathbf{0.96}\pm0.027)$ & $(0.96\pm0.027)$ & $(0.96\pm0.027)$ \\[3pt]
MGN & $(\mathbf{0.5}\pm0.49)$ & $(15\pm11)$ & $(\mathbf{98}\pm47)$ & $(1.8\pm1.7)$ & $(51\pm35)$ & $(350\pm140)$ \\[3pt]
Simulator & $(0.62\pm0.56)$ & $(18\pm11)$ & $(130\pm37)$ & $(0.94\pm0.88)$ & $(28\pm18)$ & $(190\pm59)$ \\[3pt]
CT & $(0.78\pm4.9)$ & $(10\pm5.6)$ & $(17\pm9.5)$ & $(6.4\pm5.1)$ & $(22\pm11)$ & $(29\pm20)$ \\[3pt]
RMSN & $(1.4\pm0.61)$ & $(41\pm30)$ & $(58\pm60)$ & $(1.8\pm1.1)$ & $(83\pm61)$ & $(47\pm59)$ \\[3pt]
CT-ERM & $(2\pm1.4)$ & $(9.4\pm4.8)$ & $(17\pm13)$ & $(4.6\pm2.9)$ & $(\mathbf{24}\pm14)$ & $(\mathbf{52}\pm12)$ \\[3pt]
CRN-ERM & $(3\pm14)$ & $(\mathbf{15}\pm14)$ & $(24\pm19)$ & $(5.6\pm22)$ & $(15\pm17)$ & $(-5.4\pm14)$ \\[3pt]
\bottomrule
\end{tabular}

\end{table}

\begin{table}[!htbp]\centering\footnotesize
\setlength{\tabcolsep}{3pt}
\caption{Average local dose response over ten base seeds: sliding treatment plans, $\gamma = 10$, $\tau = 5$. Units are $\mathrm{cm^3}$. Within each seed, decile-level statistics are averaged equally within ranges 0--5, 6--8 and 9. Cells give the across-seed mean $\pm$ sample standard deviation. Rows are sorted by increasing chemotherapy mean in deciles 0--5. Positive responses indicate tumor reduction. Bold means are closest to the simulator's mean. Model references are in App.~\ref{app:models}.}
\label{tab:seed-response-g10-sliding}
\begin{tabular}{lcccccc}
\toprule
 & \multicolumn{3}{c}{Chemotherapy} & \multicolumn{3}{c}{Radiotherapy} \\
\cmidrule(lr){2-4}\cmidrule(lr){5-7}
Model & Deciles 0--5 & Deciles 6--8 & Decile 9 & Deciles 0--5 & Deciles 6--8 & Decile 9 \\
\midrule
EDCT & $(-1.5\pm5.2)$ & $(2.3\pm7.2)$ & $(12\pm8.7)$ & $(-4.7\pm13)$ & $(-2.6\pm14)$ & $(1.5\pm12)$ \\[3pt]
CRN & $(-1.3\pm7.7)$ & $(-1\pm8.9)$ & $(3.1\pm9.2)$ & $(12\pm19)$ & $(13\pm16)$ & $(8.1\pm16)$ \\[3pt]
OLS1 & $(-0.66\pm0.018)$ & $(-0.66\pm0.018)$ & $(-0.66\pm0.018)$ & $(-0.044\pm0.022)$ & $(-0.044\pm0.022)$ & $(-0.044\pm0.022)$ \\[3pt]
MSM & $(0.43\pm0.014)$ & $(0.43\pm0.014)$ & $(0.43\pm0.014)$ & $(\mathbf{0.96}\pm0.027)$ & $(0.96\pm0.027)$ & $(0.96\pm0.027)$ \\[3pt]
MGN & $(\mathbf{0.5}\pm0.49)$ & $(\mathbf{15}\pm12)$ & $(\mathbf{100}\pm49)$ & $(1.8\pm1.7)$ & $(51\pm36)$ & $(360\pm140)$ \\[3pt]
Simulator & $(0.62\pm0.57)$ & $(18\pm12)$ & $(130\pm40)$ & $(0.97\pm0.89)$ & $(28\pm19)$ & $(200\pm66)$ \\[3pt]
CT & $(0.81\pm4.9)$ & $(10\pm6.3)$ & $(18\pm8.1)$ & $(6.5\pm5.2)$ & $(22\pm11)$ & $(30\pm17)$ \\[3pt]
RMSN & $(1.3\pm0.57)$ & $(41\pm30)$ & $(59\pm37)$ & $(1.6\pm1.1)$ & $(99\pm100)$ & $(\mathbf{66}\pm61)$ \\[3pt]
CRN-ERM & $(1.3\pm14)$ & $(13\pm15)$ & $(24\pm18)$ & $(4.9\pm21)$ & $(13\pm17)$ & $(-8.2\pm14)$ \\[3pt]
CT-ERM & $(2\pm1.4)$ & $(9.4\pm4.6)$ & $(17\pm14)$ & $(4.6\pm2.9)$ & $(\mathbf{24}\pm14)$ & $(54\pm13)$ \\[3pt]
\bottomrule
\end{tabular}

\end{table}

\begin{table}[!htbp]\centering\footnotesize
\setlength{\tabcolsep}{3pt}
\caption{Average local dose response over ten base seeds: sliding treatment plans, $\gamma = 4$, $\tau = 5$. Units are $\mathrm{cm^3}$. Within each seed, decile-level statistics are averaged equally within ranges 0--5, 6--8 and 9. Cells give the across-seed mean $\pm$ sample standard deviation. Rows are sorted by increasing chemotherapy mean in deciles 0--5. Positive responses indicate tumor reduction. Bold means are closest to the simulator's mean. Model references are in App.~\ref{app:models}.}
\label{tab:seed-response-g4-sliding}
\begin{tabular}{lcccccc}
\toprule
 & \multicolumn{3}{c}{Chemotherapy} & \multicolumn{3}{c}{Radiotherapy} \\
\cmidrule(lr){2-4}\cmidrule(lr){5-7}
Model & Deciles 0--5 & Deciles 6--8 & Decile 9 & Deciles 0--5 & Deciles 6--8 & Decile 9 \\
\midrule
CRN-ERM & $(-5.8\pm11)$ & $(-5.1\pm12)$ & $(20\pm19)$ & $(25\pm27)$ & $(27\pm30)$ & $(30\pm36)$ \\[3pt]
CRN & $(-3.9\pm14)$ & $(-3.9\pm15)$ & $(8.6\pm16)$ & $(5.8\pm30)$ & $(5.4\pm32)$ & $(-3.8\pm42)$ \\[3pt]
OLS1 & $(-0.089\pm0.0058)$ & $(-0.089\pm0.0058)$ & $(-0.089\pm0.0058)$ & $(-0.026\pm0.0073)$ & $(-0.026\pm0.0073)$ & $(-0.026\pm0.0073)$ \\[3pt]
EDCT & $(-0.036\pm0.44)$ & $(0.18\pm0.67)$ & $(12\pm5.1)$ & $(-1.2\pm5.3)$ & $(-1.6\pm7.3)$ & $(3.8\pm12)$ \\[3pt]
CT-ERM & $(0.0016\pm0.054)$ & $(0.86\pm1.1)$ & $(16\pm7.3)$ & $(0.14\pm0.33)$ & $(1.6\pm1.3)$ & $(25\pm11)$ \\[3pt]
CT & $(0.014\pm0.21)$ & $(0.85\pm1.6)$ & $(19\pm9.9)$ & $(0.16\pm0.81)$ & $(\mathbf{2.3}\pm4.6)$ & $(\mathbf{35}\pm23)$ \\[3pt]
MGN & $(\mathbf{0.03}\pm0.027)$ & $(\mathbf{1.8}\pm1.3)$ & $(42\pm19)$ & $(\mathbf{0.05}\pm0.049)$ & $(3\pm2.2)$ & $(76\pm37)$ \\[3pt]
Simulator & $(0.031\pm0.024)$ & $(1.7\pm1)$ & $(35\pm12)$ & $(0.046\pm0.037)$ & $(2.1\pm1.3)$ & $(42\pm16)$ \\[3pt]
MSM & $(0.11\pm0.0072)$ & $(0.11\pm0.0072)$ & $(0.11\pm0.0072)$ & $(0.19\pm0.0087)$ & $(0.19\pm0.0087)$ & $(0.19\pm0.0087)$ \\[3pt]
RMSN & $(2.8\pm1.4)$ & $(4.6\pm1.8)$ & $(\mathbf{36}\pm11)$ & $(0.33\pm0.54)$ & $(3.1\pm1.7)$ & $(30\pm15)$ \\[3pt]
\bottomrule
\end{tabular}

\end{table}

\begin{table}[!htbp]\centering\footnotesize
\setlength{\tabcolsep}{3pt}
\caption{Average local dose response over ten base seeds: random treatment plans, $\gamma = 4$, $\tau = 5$. Units are $\mathrm{cm^3}$. Within each seed, decile-level statistics are averaged equally within ranges 0--5, 6--8 and 9. Cells give the across-seed mean $\pm$ sample standard deviation. Rows are sorted by increasing chemotherapy mean in deciles 0--5. Positive responses indicate tumor reduction. Bold means are closest to the simulator's mean. Model references are in App.~\ref{app:models}.}
\label{tab:seed-response-g4-random}
\begin{tabular}{lcccccc}
\toprule
 & \multicolumn{3}{c}{Chemotherapy} & \multicolumn{3}{c}{Radiotherapy} \\
\cmidrule(lr){2-4}\cmidrule(lr){5-7}
Model & Deciles 0--5 & Deciles 6--8 & Decile 9 & Deciles 0--5 & Deciles 6--8 & Decile 9 \\
\midrule
CRN-ERM & $(-5.3\pm12)$ & $(-4.7\pm14)$ & $(20\pm20)$ & $(26\pm29)$ & $(28\pm32)$ & $(29\pm35)$ \\[3pt]
CRN & $(-5.1\pm13)$ & $(-5.2\pm14)$ & $(5.6\pm14)$ & $(1.6\pm32)$ & $(1.2\pm34)$ & $(-8.1\pm44)$ \\[3pt]
OLS1 & $(-0.089\pm0.0058)$ & $(-0.089\pm0.0058)$ & $(-0.089\pm0.0058)$ & $(-0.026\pm0.0073)$ & $(-0.026\pm0.0073)$ & $(-0.026\pm0.0073)$ \\[3pt]
EDCT & $(-0.037\pm0.44)$ & $(0.17\pm0.67)$ & $(12\pm4.9)$ & $(-1.2\pm5.3)$ & $(-1.6\pm7.2)$ & $(3.4\pm13)$ \\[3pt]
CT-ERM & $(0.0083\pm0.048)$ & $(0.9\pm1.1)$ & $(16\pm7.3)$ & $(0.15\pm0.33)$ & $(1.6\pm1.4)$ & $(24\pm11)$ \\[3pt]
CT & $(0.012\pm0.21)$ & $(0.87\pm1.7)$ & $(19\pm10)$ & $(0.16\pm0.81)$ & $(\mathbf{2.3}\pm4.8)$ & $(34\pm21)$ \\[3pt]
MGN & $(\mathbf{0.033}\pm0.032)$ & $(\mathbf{1.8}\pm1.3)$ & $(42\pm19)$ & $(\mathbf{0.055}\pm0.058)$ & $(3\pm2.3)$ & $(76\pm37)$ \\[3pt]
Simulator & $(0.033\pm0.027)$ & $(1.7\pm1)$ & $(35\pm12)$ & $(0.049\pm0.042)$ & $(2.1\pm1.4)$ & $(41\pm16)$ \\[3pt]
MSM & $(0.11\pm0.0072)$ & $(0.11\pm0.0072)$ & $(0.11\pm0.0072)$ & $(0.19\pm0.0087)$ & $(0.19\pm0.0087)$ & $(0.19\pm0.0087)$ \\[3pt]
RMSN & $(2.4\pm1)$ & $(4.1\pm0.96)$ & $(\mathbf{40}\pm14)$ & $(0.65\pm1.1)$ & $(3.5\pm2.1)$ & $(\mathbf{38}\pm16)$ \\[3pt]
\bottomrule
\end{tabular}

\end{table}

\clearpage
\section{Evaluation settings in prior studies}
\label{app:prior-settings}

The studies in Tables~\ref{tab:prior-designs} and \ref{tab:prior-reporting} evaluate
models on data from the same tumor-growth simulator, with different evaluation settings.
The tables summarize their tumor prediction comparisons. They retain each paper's horizon
labels; matching labels alone do not establish identical forecast tasks. NR means that a
setting is not explicitly reported in the cited experimental text, tables or appendices.

\paragraph{Model coverage.}
The original MSM paper \citep{robins2000} does not report a tumor evaluation;
\citet{lim2018} evaluate MSM alongside RMSN.
\citet{shalit2017} introduce TARNet and CFR and compare OLS1 and OLS2 on static
treatment tasks. Their tumor evaluations appear in \citet{huang2024}.
CRN, G-Net, CT and INSITE are evaluated on simulated tumor data in their original
papers \citep{bica2020,li2021,melnychuk2022,kacprzyk2024}.
EDCT appears in the architecture ablation of \citet{melnychuk2022}.
CRN and CT without balancing already appear in \citet{bica2020} and
\citet{melnychuk2022}, respectively; \citet{huang2024} compare both with their
balanced counterparts. We label these variants CRN-ERM and CT-ERM.

\begingroup
\footnotesize
\setlength{\tabcolsep}{3pt}
\begin{longtable}{@{}>{\raggedright\arraybackslash}p{0.14\linewidth}>{\raggedright\arraybackslash}p{0.17\linewidth}>{\raggedright\arraybackslash}p{0.26\linewidth}>{\raggedright\arraybackslash}p{0.09\linewidth}>{\raggedright\arraybackslash}p{0.26\linewidth}@{}}
\caption{Assignment strengths, horizons and evaluation plans in prior tumor prediction experiments. Assignment strength refers to $\gamma$; a single value applies to both treatments unless stated otherwise. Horizon labels follow each source.}
\label{tab:prior-designs}\\
\toprule
Study & Experiment & Assignment strength $\gamma$ & Horizon & Evaluation plans \\
\midrule
\endfirsthead
\multicolumn{5}{l}{\tablename\ \thetable\ (continued)}\\
\toprule
Study & Experiment & Assignment strength $\gamma$ & Horizon & Evaluation plans \\
\midrule
\endhead
\midrule
\endfoot
\bottomrule
\endlastfoot
\citet{lim2018} & one-step & Training: $10$ for both treatments.\newline Test: $0$--$10$ for both treatments. & $1$ & Trajectories generated by the test assignment policy. \\*
\cmidrule[0.3pt]{2-5}
 & multi-step & Training: $10$ for both treatments.\newline Four test settings:\newline $10$ for both;\newline $0$ for both;\newline chemotherapy $10$, radiotherapy $0$;\newline chemotherapy $0$, radiotherapy $10$. & $1$--$5$ & Trajectories generated by the test assignment policy. At $\gamma=0$, assignment is independent of tumor history for that treatment. \\
\midrule[0.5pt]
\citet{bica2020} & one-step & $0$--$10$ & $1$ & All four treatment combinations. \\*
\cmidrule[0.3pt]{2-5}
 & multi-step & $1$--$5$ & $2$--$6$ & Sliding treatment plans. \\*
\cmidrule[0.3pt]{2-5}
 & treatment selection & $5$ for both;\newline chemotherapy $5$, radiotherapy $0$;\newline chemotherapy $0$, radiotherapy $5$. & $3$--$7$ & Choose the treatment and its application day. \\
\midrule[0.5pt]
\citet{li2021} & Fixed regimens & $10$ & $1$--$4$ & Four fixed regimens at the end of a trajectory: neither treatment, chemotherapy only, radiotherapy only, or both. \\
\midrule[0.5pt]
\citet{melnychuk2022} & prediction comparison & $0$--$4$ & $1$; $2$--$6$ & All four combinations at one step; sliding and random treatment plans at multiple steps. \\*
\cmidrule[0.3pt]{2-5}
 & architecture ablation & $1$, $4$ & $1$, $6$ & Random treatment plans; this is the reported comparison including EDCT. \\
\midrule[0.5pt]
\citet{huang2024} & standard / short history & $1,3,10$ in main tables; $2,6,8$ in appendix tables. & $1$--$5$ / $1$--$6$\newline (main / appendix) & NR. Separate assignment-shift experiments train at $10$ and test at $1,3,8$ (main) or $2,6$ (appendix). \\*
\cmidrule[0.3pt]{2-5}
 & sequential / non-sequential & $0,1,2,3,6,8,10$ & NR & NR. Includes TARNet, CFR, OLS1 and OLS2 alongside CT, CRN and their ERM variants. \\
\midrule[0.5pt]
\citet{kacprzyk2024} & Cancer PKPD & $2$ in main comparison; $0$--$4$ in assignment sweep. & $1$--$6$; sweep at $6$ & All four combinations at one step; sliding treatment plans at multiple steps. Main Table 3 reports horizon $6$. \\
\end{longtable}
\endgroup

\paragraph{Scoring and data.}
The studies also differ in the outcomes they score, cohort sizes and repeated runs
(Table~\ref{tab:prior-reporting}). A shared metric name does not specify its
normalization or which forecast steps contribute to it. The model's input history
is a separate choice from the 15-day history used to assign treatments.

\begingroup
\footnotesize
\setlength{\tabcolsep}{3pt}
\begin{longtable}{@{}>{\raggedright\arraybackslash}p{0.14\linewidth}>{\raggedright\arraybackslash}p{0.3\linewidth}>{\raggedright\arraybackslash}p{0.3\linewidth}>{\raggedright\arraybackslash}p{0.18\linewidth}@{}}
\caption{Scoring and data reported for the experiments in Table~\ref{tab:prior-designs}. Cohort sizes are training / validation / test patients. NR means not explicitly reported in the cited experimental text, tables or appendices.}
\label{tab:prior-reporting}\\
\toprule
Study & Error and scored outcomes & Cohorts and model history & Repeated runs \\
\midrule
\endfirsthead
\multicolumn{4}{l}{\tablename\ \thetable\ (continued)}\\
\toprule
Study & Error and scored outcomes & Cohorts and model history & Repeated runs \\
\midrule
\endhead
\midrule
\endfoot
\bottomrule
\endlastfoot
\citet{lim2018} & Metric: nRMSE (\%).\newline Normalization: $1{,}150\,\mathrm{cm^3}$.\newline Scoring: historical/future split NR. & $10{,}000/1{,}000/1{,}000$; MSM combines training and validation. Recurrent history; fixed input length NR. & Count and uncertainty: NR. \\
\midrule[0.5pt]
\citet{bica2020} & Metric: nRMSE (\%).\newline Normalization: $1{,}150\,\mathrm{cm^3}$.\newline Scoring: historical/future split NR. Also reports treatment/timing selection scores. & $10{,}000/1{,}000/1{,}000$ per assignment setting; evaluate each patient/time origin. Fixed input length NR. & Count and uncertainty: NR. \\
\midrule[0.5pt]
\citet{li2021} & Metric: nRMSE (\%).\newline Normalization: $1{,}150\,\mathrm{cm^3}$.\newline Scoring: final four future outcomes, separately by forecast step and overall. & $10{,}000/1{,}000/1{,}000$; condition on prior observations. Fixed input length NR. & Tumor comparison count and uncertainty: NR. \\
\midrule[0.5pt]
\citet{melnychuk2022} & Metric: nRMSE.\newline Normalization: $1{,}150\,\mathrm{cm^3}$.\newline Scoring: future outcomes only, separately by forecast step. & $10{,}000/1{,}000/1{,}000$; rolling forecast origins. Fixed model context length NR. & Five independently simulated datasets; mean and SD. \\
\midrule[0.5pt]
\citet{huang2024} & Metric: RMSE.\newline Normalization: NR.\newline Scoring: separately by forecast horizon in standard tables; exact scoring mask NR. & Cohort sizes NR. Short-history truncation is denoted $K$; numeric $K$ NR. & Five runs; mean and SD. Data regeneration per run NR. \\
\midrule[0.5pt]
\citet{kacprzyk2024} & Metric: nRMSE.\newline Normalization: $1{,}150\,\mathrm{cm^3}$.\newline Scoring: separately by forecast horizon. & Default $1{,}000/100/100$; every patient/time origin. INSITE adapts coefficients to patient history; fixed context length NR. & New datasets per run. Main comparison: $10$ runs; sweep: $5$. Mean and 95\% CI. \\
\end{longtable}
\endgroup

\paragraph{Our evaluation.}
Our implementation builds on the simulator and dataset-generation pipeline released
with Causal Transformer \citep{melnychuk2022}. We retain its sliding and random
treatment-plan constructions, cohort sizes and training-data scaling. We report nRMSE
relative to approximately $1{,}150\,\mathrm{cm^3}$ and score future outcomes rather
than the observed history. We use the released code's default projection length of
five days and report aggregate prediction error at the final forecast step
(\S\ref{sec:setup}); \citet{melnychuk2022} report errors separately at multiple
forecast steps. Our main assignment strength, $\gamma=10$, is outside their
reported range of $0$--$4$, while our $\gamma=4$ results overlap with it.
Our MGN implementation uses the last 16 days without patient type, whereas the
released CT implementation uses the available history and patient type.
App.~\ref{app:models} specifies history inputs for every model.
App.~\ref{app:mgn} describes patient-type inputs.

\end{document}